\documentclass[runningheads]{llncs}
\usepackage{graphicx}
\usepackage{comment}
\usepackage{amsmath,amssymb} 
\usepackage{color}

\usepackage{multirow}
\usepackage{amssymb,amsmath,graphicx,amsfonts,euscript}
\usepackage{amsmath}
\usepackage{mathtools}
\usepackage[flushleft]{threeparttable}
\usepackage{makecell,booktabs}
\usepackage{algorithm}
\usepackage{algpseudocode}
\usepackage{color}
\usepackage{caption}
\usepackage{subcaption}
\usepackage{enumitem}
\setlist{nolistsep}

\newcommand{\equalref}{\operatornamewithlimits{=}}

\usepackage[width=122mm,left=12mm,paperwidth=146mm,height=193mm,top=12mm,paperheight=217mm]{geometry}
\newcommand{\myfontsize}{\normalsize}

\begin{document}
\pagestyle{headings}
\mainmatter
\def\ECCVSubNumber{866}  

\title{Robust Neural Networks inspired by Strong Stability Preserving Runge-Kutta methods} 

\titlerunning{Robust Neural Networks inspired by SSP Runge-Kutta methods}
%
\author{Byungjoo Kim\inst{1}\thanks{Equal Contribution, $^\dagger$ Corresponding Author.} \and
Bryce Chudomelka\inst{2}$^\star$ \and
Jinyoung Park\inst{1} \and\\
Jaewoo Kang\inst{1}$^\dagger$ \and
Youngjoon Hong\inst{2} \and
Hyunwoo J. Kim\inst{1}$^\dagger$
}

\authorrunning{B. Kim et al.}
%
\institute{Department of Computer Science, Korea University, Seoul, Republic of Korea \\
\email{\{byung4329,lpmn678,kangj,hyunwoojkim\}@korea.ac.kr}\\
\and
Department of Mathematics and Statistics, San Diego State University, San Diego, California, USA \\
\email{\{bchudomelka,yhong2\}@sdsu.edu}}
\maketitle

\begin{abstract}
Deep neural networks have achieved state-of-the-art performance in a variety of fields.
Recent works observe that a class of widely used neural networks can be viewed as the Euler method of numerical discretization.
From the numerical discretization perspective, Strong Stability Preserving (SSP) methods are more advanced techniques than the explicit Euler method that produce both accurate and stable solutions.
Motivated by the SSP property and a generalized Runge-Kutta method, we proposed Strong Stability Preserving networks (SSP networks) which improve robustness against adversarial attacks.
We empirically demonstrate that the proposed networks improve the robustness against adversarial examples without any defensive methods.
Further, the SSP networks are complementary with a state-of-the-art adversarial training scheme.
Lastly, our experiments show that SSP networks suppress the blow-up of adversarial perturbations.
Our results open up a way to study robust architectures of neural networks leveraging rich knowledge from numerical discretization literature.
\end{abstract}

\section{Introduction}
\begin{figure}[t]
    \centering
    \begin{subfigure}[b]{0.3\columnwidth}
        \centering
        \includegraphics[width=\textwidth]{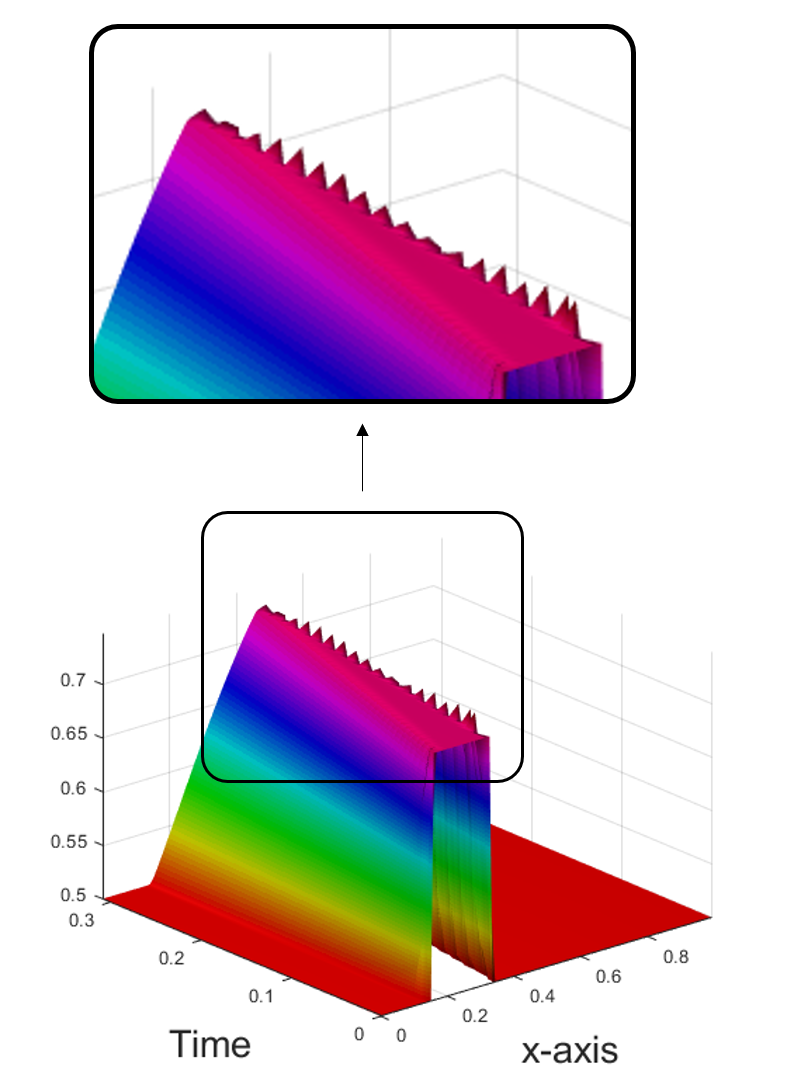}
        \caption{Euler method}
        \label{subfig:euler}
    \end{subfigure}%
    \begin{subfigure}[b]{0.3\columnwidth}
        \centering
        \includegraphics[width=\textwidth]{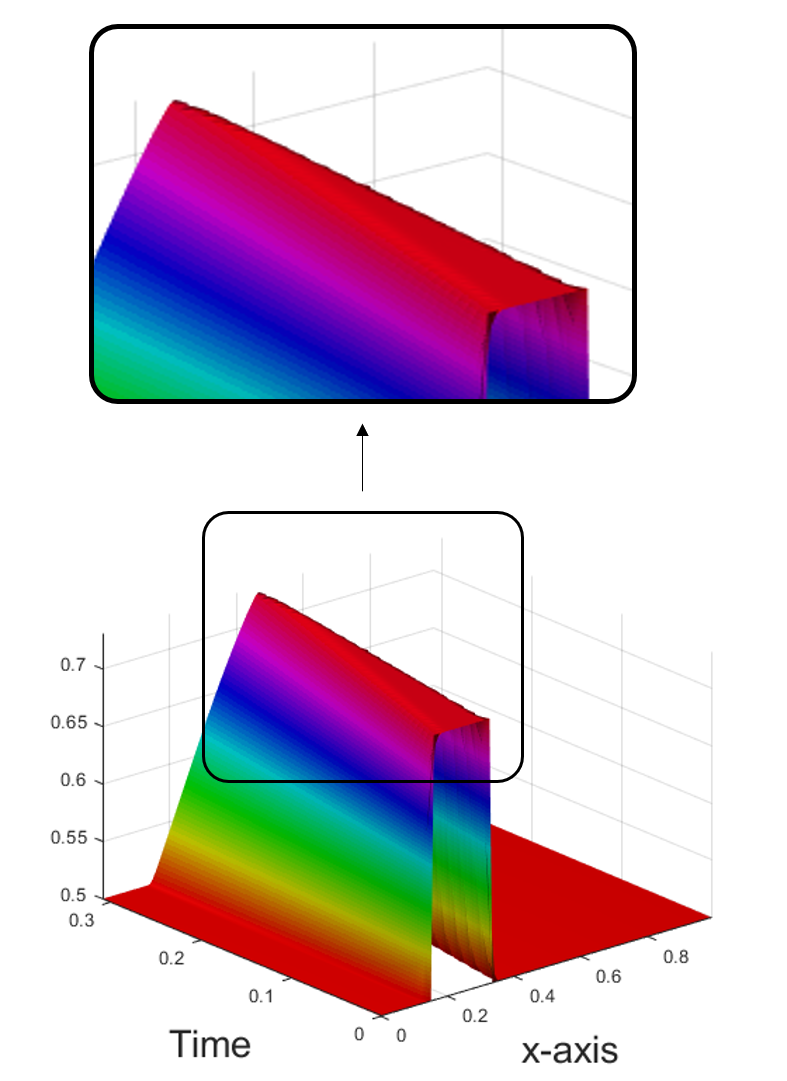}
        \caption{3rd order SSP}
        \label{subfig:ssp3}
    \end{subfigure}
    \caption{
    We illustrate the difference between a forward Euler discretization and a  third-order SSP discretization applied to the inviscid Burgers' solution.
    After computing numerical solutions, the solutions are filtered through the sigmoid function as an activation function.
    Evidently, in (a) the Euler scheme, i.e., a \textit{ResBlock}, produces notable numerical errors while the SSP3 discretization in (b) shows a stable  numerical approximation. 
    For more details, please see the supplement.
    }
    \label{fig:title_figure}
\end{figure}
Recent progress in deep learning has shown promising results in various research areas, such as computer vision, natural language processing and recommendation systems.
In particular, on the ImageNet classification task \cite{krizhevsky2009learning}, deep neural networks show state-of-the-art performance, e.g., residual networks (ResNet), which outperform humans in image classification \cite{he2016deep}.
Despite the success, deep neural networks often suffer from the lack of robustness against adversarial attacks \cite{szegedy2013intriguing}.
ResNet, which is a widely used base network, also suffers from adversarial attacks which necessitates a more fundamental understanding of the architecture at hand.

One interesting interpretation of the ResNet architecture is that of the explicit Euler discretization scheme, i.e., $x(t_{k+1})=x(t_k) + F(x(t_k))$, because it allows one to view neural networks as numerical methods.
The explicit Euler method is one of the simplest first-order numerical schemes but often leads to large numerical errors due to its low order.
Thus, we would expect that applying advanced numerical discretizations would produce a more accurate numerical solution than the Euler method, such as an explicit high-order Runge-Kutta method.
However, an arbitrary explicit high-order Runge-Kutta method can pose a stability problem if the numerical solution becomes unstable~\cite{shu1988total}.
To tackle this issue, \cite{gottlieb1998total} and \cite{shu1988total} introduce the notion of total variation diminishing (TVD); also called Strong Stability Preserving (SSP) methods.
The strong stability preserving approach produces a more accurate solution of the differential equation than the Euler method.
We would expect to obtain a more accurate solution of the underlying function with non-smooth initial data (shocks) compared to the Euler method without notable numerical errors, see Figure
\ref{fig:title_figure}.
This phenomenon is directly related to the problem of adversarial attack and robustness of neural networks~\cite{szegedy2013intriguing}.

Motivated by the advanced numerical discretization schemes, we propose novel network architectures with the SSP property that address robustness; SSP networks (SSPNets). 
The use of the SSP property consistently demonstrates that all of our proposed architectures outperform ResNet in terms of robustness.
SSP architectural blocks do not increase the amount of model parameters compared to ResNet, can be easily implemented, and realized by a convex combination of existing ResNet modules.
The parameters used in SSP blocks are {\it mathematically derived coefficients} from the advanced numerical discretization methods.
In addition, starting from an explicit Runge-Kutta method with the SSP property,
we propose novel Adaptive Runge-Kutta blocks with {\it learned coefficients} obtained by training.
\textcolor{black}{With these learned coefficients, we are able to improve robustness while retaining the natural accuracy of ResNet.}

The simple architectural change, SSPNets, improve robustness and are complementary with adversarial training, which is the \textit{de facto} state-of-the-art defensive methodology.
Our \textbf{contributions} are summarized as follows:
\begin{itemize}
    \item We propose multiple novel architectural blocks motivated by the Strong Stability Preserving explicit higher-order numerical discretization method.
    \item We demonstrate empirically that these proposed blocks improve the robustness of Strong Stability Preserving networks consistently; against adversarial examples and without any defensive methods.
    \textcolor{black}{\item We further improve on robustness with a novel adaptive architectural block motivated by a generalized Runge-Kutta method and the SSP property.}
    \item Last but not least, we show that Strong Stability Preserving Networks suppress the blow-up of adversarial perturbations added to inputs.
\end{itemize}

\section{Background and Related Work}
\subsection{Neural Networks and Differential Equations}
Neural networks such as ResNet \cite{he2016deep}, PolyNet \cite{zhang2017polynet} and recurrent neural networks share a common operation represented as $x_{t+1} = x_t + F(x_t;\Theta_t)$.
Interestingly, a sequence of the operations (or equivalently the network architectures) can be interpreted as an explicit Euler method for numerical discretization \cite{chen2018neural,ciccone2018nais,lu2017beyond,rubanova2019latent,ruthotto2018deep}.
For instance, ResNet can be written mathematically as
\begin{equation}
\label{eq:resnet}
\begin{split}
    & x_0 = x,  \\
    & x_{k+1} = x_k + F(x_k; \Theta_k), \quad k \in \{ 0,1,\dots,A-1 \},   \\
    & \hat{y} = f(x_A),
\end{split}
\end{equation}
where $A$ denotes the number of layers in the network.

If we multiply the function $F$ by $\Delta t$, i.e., $x_{k+1}=x_k + \Delta t F(x_k;\Theta_k)$, then ResNet can be seen as the explicit Euler numerical scheme discretization with an initial condition, $x(0)$, to solve the initial value problem given as
\begin{equation}
\label{eq:diffeq}
\begin{split}
    & x(0) = x ,\\
    & \frac{dx(t)}{dt} = F(x(t);\Theta(t)),   \\
    & \hat{y}=f(x(A)).
\end{split}
\end{equation}

The explicit Euler method is the simplest Runge-Kutta method and often suffers from low accuracy because it is a first-order method. 
In this regard, higher-order numerical methods are natural candidates to obtain a more precise numerical solution, but the higher accuracy from higher-order methods may come with the cost of instability, e.g.,  poor convergence behaviour on stiff differential equations compared to the first order Euler method~\cite{Butcher87textbook}. Therefore, it is important to understand the trade-off between accuracy and stability when considering a numerical method.

Recently, some network architectures inspired by the computational similarity between ResNet and Euler discretization have been proposed, e.g., NeuralODE and FFJORD \cite{chen2018neural,grathwohl2018ffjord}.
Unlike ResNet, which requires the discretization of observation/emission intervals to be represented by 
a finite number of hidden layers, NeuralODE and FFJORD use 
numerical discretization methods in the forward propagation to define continuous-depth and continuous-time latent variable models. These require ODE solvers for training and inference, unlike our implementation of SSP networks. Since we changed only computational graphs and coefficients based on the numerical discretization theory, our methods perform the standard forward/backward propagation in the discrete space as ResNet.

Another approach to design new blocks/layers of neural networks is to make them have operations similar to advanced numerical discretization techniques that possess desirable properties~\cite{lu2017beyond,ruthotto2018deep}. 
From the partial differential equation perspective, analysis on numerical stability of conventional residual connections lead to the development of new architectures: parabolic/hyperbolic CNNs to achieve better stability as parabolic/hyperbolic PDEs \cite{ruthotto2018deep}.
The models use theoretical assumptions on the function to achieve stability with a positive semi-definite Jacobian of the function resulting in constraints on convolutional kernels; alternatively, our networks do not require such constraints.

\subsection{Robust Machine Learning and Adversarial Attacks}
Stability and robustness of neural networks have been studied in the context of adversarial attacks after the success of deep learning \cite{ben2009robust,szegedy2013intriguing,xu2009robustness}.
Gradient-based adversarial attacks create adversarial examples solving optimization problems.
One example is the maximization of loss against ground truth labels within a small ball, e.g., $\max_{\delta} \mathcal{L}(h_{\theta}(x+\delta), y),\textit{ s.t. }\|\delta\|_\infty \le \epsilon$, where $h_{\theta}$ is a model parameterized by $\theta$, $x$, $y$ are the input (natural sample) and its target label respectively, and $\mathcal{L}$ is a loss function. 
The simplest procedure to approximate the solution is to use the fast gradient sign method (FGSM)  \cite{goodfellow2014explaining}.
It can be seen as an optimal solution to a linearized loss function, i.e., 
$\arg\max_{\|v \|_{\infty} \leq \alpha} v^{T}\nabla_\delta \mathcal{L}(h_{\theta}(x+\delta), y) = \alpha \cdot \text{sign}(\nabla_\delta \mathcal{L}(h_{\theta}(x+\delta), y))$.
Furthermore, the FGSM can be more powerful when it is used with iterative methods such as the projected gradient descent (PGD). PGD has been used in both untargeted and targeted attacks \cite{madry2017towards,carlini2017towards}.

One of the early attempts to defend against adversarial attacks is adversarial training using FGSM, a single-step method \cite{goodfellow2014explaining}.
After that, various defensive techniques have been proposed \cite{buckman2018thermometer,papernot2016distillation,samangouei2018defensegan,song2018pixeldefend,tsipras2018robustness}. Many of them were defeated by iterative attack methods \cite{carlini2017towards} and Backward Pass Differentiable Approximation \cite{athalye2018obfuscated}.
Adversarial training with stronger multi-step attack methods is still promising and shows state-of-the-art performance  \cite{madry2017towards,wang2019on,xie2019feature}. More recently, \textit{provably} robust neural networks have been successfully trained by minimizing the lower bound of risk based on convex duality and convex relaxation \cite{wong2017provable,wong2018scaling}. 
Most adversarial training methods above assume that attack methods are known \textit{a priori}, i.e., a \textit{white-box} attack, and generate augmented samples using the attacks. Another defensive technique is to alleviate the effect of perturbation by augmentation and reconstruction \cite{raff2019bart,yang2019me}, or denoising \cite{xie2019feature}. These methods alongside adversarial training achieved comparable robustness. Similarly, in this work we will introduce our approach and evaluate it with adversarial training.

\section{Strong Stability Preserving Networks}

\textcolor{black}{
In this section, we introduce the mathematical framework for the Strong Stability Preserving property and describe how to implement SSP blocks with \textit{mathematically derived coefficients}. 
Next, we provide a variance analysis to compare high-order Runge-Kutta blocks with residual blocks.
Lastly, we introduce adaptive Runge-Kutta blocks with \textit{learnable coefficients} which possess the SSP property.
}
\subsection{Motivation of strong stability preserving method}

Our objective is to solve the \textit{non-autonomous} differential equation given as
\begin{equation}
\label{eq:initial_value_problem}
    \frac{\partial u}{\partial t} = L(u(t), t), \quad t \in [t_0,...,t_N],
\end{equation}
where  $t_0, t_N$ are the initial and terminal time state respectively; a non-autonomous system permits a time varying solution, e.g., the learned function varies as the depth of the network increases. The function $L$ is a linear (or nonlinear) function and $u(t_0)$ is given by the initial condition.
The objective is to figure out the terminal state of the function $u$, i.e., $u(t_N)$.

A general high-order Runge-Kutta time discretization for solving the initial value problem \eqref{eq:initial_value_problem} introduced in  \cite{shu1988efficient} is given as
\begin{equation}
\label{eq:runge-kutta}
 \begin{split}
     & u^{(0)} = u^n ,   \\
     & u^{(i)} = \sum^{i-1}_{k=0} \left( \alpha_{i,k} u^{(k)} + \Delta t \beta_{i,k} L(u^{(k)}) \right), \quad i\in\{1,\cdots,m\},\\
     & u^{n+1} = u^{(m)},
    \end{split}
\end{equation}
where $\sum^{i-1}_{k=0} \alpha_{i,k} = 1$ and $\alpha_{i,k} \geq 0$.
For example, if $m=1$, it becomes the first-order Euler method as in Equation \eqref{eq:resnet} with $\alpha_{1,0}=\beta_{1,0}=1$. 

Shu et al. \cite{shu1988total,shu1988efficient} propose a TVD time discretization method that is called the SSP time discretization method; for more discussion on the TVD method, we refer the reader to \cite{Harten83,Harten87}.
The procedure of TVD time discretization is to take the high-order method to decrease the local truncation error and maintain the stability under a suitable restriction on the time step. 
While applying the TVD scheme into the explicit high-order Runge-Kutta methods, there needs the assumption to hold it: \textit{The first-order Euler method in time is strongly stable under a certain (semi) norm when the time step $\Delta t$ is suitably restricted} \cite{gottlieb2001Siam}.
More precisely, if we assume that the forward Euler time discretization is stable under a certain norm, the SSP methods find a higher-order time discretization that maintains strong stability for the same norm; improving accuracy.

Followed by this assumption, for a sufficiently small time step known as Courant-Friedrichs-Lewy (CFL) condition $\Delta t \le \Delta t_{CFL}$, the total variation semi-norm of the numerical scheme does not increase in time, that is,
\begin{equation}
\label{eq:TV}
    TV(u^{n+1}) \le TV(u^n),
\end{equation}
where the total variation is defined by
\begin{equation}
\label{eq:total_variation}
    TV(u^n) := \sum_j |u_{j+1}^n - u_j^n|,
\end{equation}
where $j$ is the spatial discretization.
The explicit high-order Runge-Kutta discretization with the SSP property maintains a higher order accuracy with a modified CFL condition $\Delta t \le c\Delta t_{CFL}$. 
In other words, the high-order SSP Runge-Kutta scheme improves accuracy while retaining its stability. This has been theoretically studied by the following Lemma \ref{lem:rk_to_ssp}.

\begin{lemma}
If the forward Euler method is strongly stable under the CFL condition, i.e. $||u^n+\Delta tL(u^n)|| \le ||u^n||$, then the Runge-Kutta method possesses SSP, $||u^{n+1}|| \le ||u^n||$, provided that $\Delta t \le c\Delta t_{CFL}.$
\label{lem:rk_to_ssp}
\end{lemma}

We provide a sketch of the proof of Lemma \ref{lem:rk_to_ssp} in the supplement. The full proof of the Lemma \ref{lem:rk_to_ssp} can be found in \cite{shu1988efficient}. 
Following this representation, we can figure out the specific coefficients $\alpha_{i,k}$ and $\beta_{i,k}$ in equation \eqref{eq:runge-kutta}.
In particular, the second and third order nonlinear SSP Runge-Kutta method was studied in \cite{shu1988efficient}.

\begin{lemma}  
An optimal second-order SSP Runge-Kutta method is given by,
\begin{equation}
\label{eq:ssp2}
\begin{split}
    & u^{(1)} = u^n + \Delta t L(u^n),    \\
    & u^{n+1} = \frac{1}{2}u^n + \frac{1}{2}u^{(1)}+\frac{1}{2}\Delta tL(u^{(1)}),
\end{split}
\end{equation}
with a CFL coefficient $c=1$.
In addition, an optimal third-order SSP Runge-Kutta method is of the form
\begin{equation}
\label{eq:ssp3}
\begin{split}
    & u^{(1)} = u^n + \Delta tL(u^n),    \\
    & u^{(2)} = \frac{3}{4}u^n + \frac{1}{4}u^{(1)}+\frac{1}{4}\Delta tL(u^{(1)}),   \\
    & u^{n+1} = \frac{1}{3}u^n + \frac{2}{3} u^{(2)} + \frac{2}{3} \Delta tL(u^{(2)}),
\end{split}
\end{equation}
with a CFL coefficient $c=1$.
\label{lem:ssp2_definition}
\end{lemma}

A sketch of the proof for Lemma \ref{lem:ssp2_definition} can be found in the supplement and for the detailed proof, we refer the reader to  \cite{gottlieb1998total,gottlieb2001Siam,shu1988efficient}.

\subsection{Strong Stability Preserving Networks}
\begin{figure}[t]
    \centering
    \begin{subfigure}[t]{0.245\textwidth}
        \centering
        \raisebox{0.1\height}{\includegraphics[width=\textwidth]{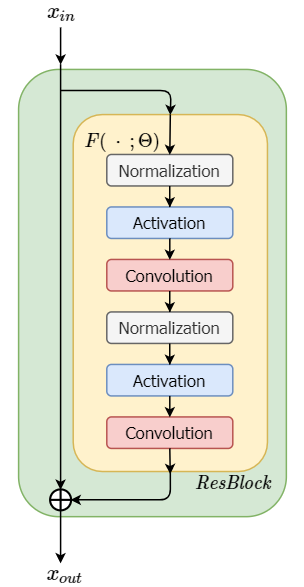}}
        \caption{}
    \end{subfigure}%
    \begin{subfigure}[t]{0.245\textwidth}
        \centering
        \raisebox{0.2\height}{\includegraphics[width=\textwidth]{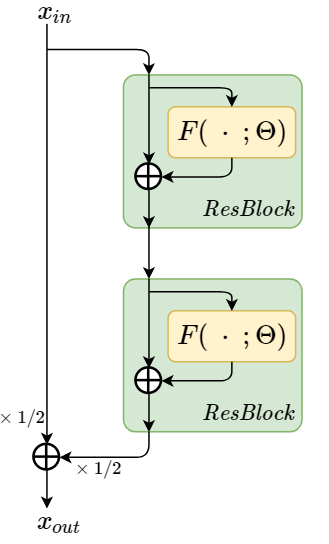}}
        \caption{}
    \end{subfigure}%
    \begin{subfigure}[t]{0.245\textwidth}
        \centering
        \includegraphics[width=\textwidth]{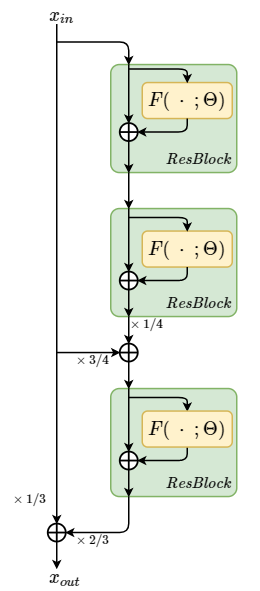}
        \caption{}
    \end{subfigure}%
    \begin{subfigure}[t]{0.245\textwidth}
        \centering
        \raisebox{0.05\height}{\includegraphics[width=\textwidth]{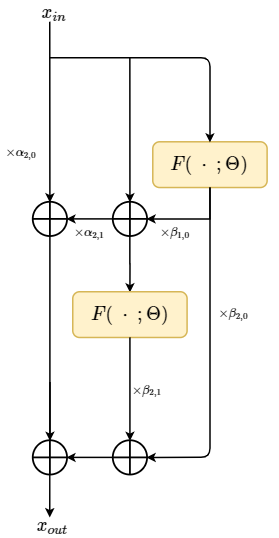}}
        \caption{}
    \end{subfigure}%
    \caption{\myfontsize
    Network modules with \textit{ResBlock} and SSP blocks.
     (a): \textit{ResBlock}. 
     (b): \textit{SSP2-block}
     (c): \textit{SSP3-block}, 
     (d): \textit{ArkBlock}
    }
    \label{fig:ssp_schematic}
\end{figure}
Next, we show how to incorporate the explicit SSP Runge-Kutta method into neural networks.
Equation \eqref{eq:ssp2} and \eqref{eq:ssp3} can be implemented with standard residual blocks and simple operations, as shown in Figure \ref{fig:ssp_schematic}.

Let \textit{ResBlock} denote a standard residual block written as
$\textit{ResBlock}(x(t_k);\Theta(t_k))$
$= x(t_k)+F(x(t_k);\Theta(t_k)),$
where $\Theta(t_k)$ are the parameters of \textit{ResBlock}($\cdot$;$\Theta (t_k)$).
The function $F$ is typically composed of two or three sets of normalization, activation, and convolutional layers, e.g., Figure \ref{fig:ssp_schematic} and \cite{he2016deep,he2016identity}.
When the numbers of input and output channels differ, we use the expansive residual block \textit{ResBlock-E}; this can be implemented with a $1\times 1$ convolutional filter to expand the number of channels.


Using the standard modules in ResNet (\textit{ResBlock} and \textit{ResBlock-E}), SSPNets can be constructed.
First, SSP blocks can be implemented using linear combinations of \textit{ResBlock}s.
As the Euler method interpretation of ResNet requires $\Delta t=1$, we assume $\Delta t=1$ in Equation \eqref{eq:ssp2}, then the \textit{SSP2-block} is given by,
\begin{equation}
\label{eq:ssp2_implement}
\begin{split}
    & x({t_{k+\frac{1}{2}}}) = \underbrace{x(t_k) + F(x(t_k);\Theta(t_k) )}_{\textit{ResBlock}\left(x\left(t_k\right);\Theta\left(t_k\right)\right)}, \\
    & x({t_{k+1}}) = \frac{1}{2} x(t_k) + \underbrace{\frac{1}{2} x({t_{k+\frac{1}{2}}}) + \frac{1}{2} F\left(x(t_{k+\frac{1}{2}});\Theta(t_k) \right)}_{\frac{1}{2} \textit{ResBlock}\left(x\left(t_{k+1/2}\right);\Theta\left(t_k\right)\right)}.
\end{split}
\end{equation}
Similarly, the third order SSP in Equation \eqref{eq:ssp3} (\textit{SSP3-block}) is written as
\begin{equation}
\label{eq:ssp3_implement}
\begin{split}
    & x({t_{k+\frac{1}{3}}}) = \underbrace{x(t_k) + F\left(x(t_k);\Theta(t_k)\right) }_{ \textit{ResBlock}\left(x\left(t_k\right);\Theta\left(t_k\right)\right) },    \\
    & x({t_{k+\frac{2}{3}}}) = \frac{3}{4}x(t_k) + \underbrace{\frac{1}{4} x({t_{k+\frac{1}{3}}}) + \frac{1}{4}F \left(x({t_{k+\frac{1}{3}}});\Theta(t_k)\right)}_{\frac{1}{4} \textit{ResBlock}\left(x\left(t_{k+1/3}\right);\Theta\left(t_k\right)\right)},   \\
    & x(t_{k+1}) = \frac{1}{3}x(t_k) + \underbrace{\frac{2}{3} x({t_{k+\frac{2}{3}}}) + \frac{2}{3} F\left(x({t_{k+\frac{2}{3}}});\Theta(t_k)\right)}_{\frac{2}{3} \textit{ResBlock}\left(x\left(t_{k+2/3}\right);\Theta\left(t_k\right)\right)}.
\end{split}
\end{equation}
The SSP block schematic is presented in Figure \ref{fig:ssp_schematic} and SSP blocks are only used when the number of channels does not change.



The explicit SSP Runge-Kutta methods in Equation \eqref{eq:ssp2} and \eqref{eq:ssp3} use the same function $L$ multiple times.
Similarly, SSP blocks in Equation \eqref{eq:ssp2_implement} and \eqref{eq:ssp3_implement} apply the same \textit{ResBlock} multiple times.
Using the same \textit{ResBlock} multiple times can be viewed as parameter sharing, which is a kind of regularization.
In other words, without increasing the number of parameters, a SSP block implementation improves the robustness of neural networks by utilizing higher-order schemes.

\noindent\textbf{Midpoint Runge-Kutta Second-Order Methods.}\quad
For contrast, one may ask whether or not the stability preserving properties are key to the robustness against adversarial perturbation.
We address this important question by training another network that utilizes a second-order midpoint Runge-Kutta method (mid-RK2) which does not have the strong stability preserving property \cite{Butcher87textbook,gottlieb2001Siam}.
Recall that this method is implemented numerically as
\begin{equation}
\label{eq:MRK2}
x(t_{k+1}) = x(t_k)+F\left(x(t_k)+\frac{1}{2}F(x(t_k);\Theta(t_k));\Theta(t_k)\right),
\end{equation}
and does not have the SSP property.
This network will provide a comparison of numerical discretization methods with regard to stability in attacked accuracy.

\noindent\textbf{Variance Analysis of SSP networks.}\quad
We analyze the variance increase of SSP blocks following previous works~\cite{he2015delving,zhang2018residual}, which compare the variance of input and output of functional modules.
Next, we show that SSP blocks suppress the variance increase compared to \textit{ResBlock}; as well as comparing the variance of the midpoint Runge-Kutta second-order numerical method for further justification.
\begin{lemma}
\label{lem:var}
If $\texttt{Var}[F(x)] = \texttt{Var}[x]$, $\texttt{Cov}[x,F(y)] = 0$ then the variance increases by
\begin{equation}
\label{eq:variance}
\begin{split}
    & \texttt{Var}[ResBlock(x)] \;\;= 2\texttt{Var}[x], \;\;\;\;\;\; \texttt{Var}[\textit{mid-RK2(x)}] \;\;\;\:= \frac{9}{4}\texttt{Var}[x], \\
    & \texttt{Var}[\textit{SSP2-Block(x)}] = \frac{7}{4}\texttt{Var}[x], \;\;\;\;\; \texttt{Var}[\textit{SSP3-Block(x)}] = \frac{29}{18}\texttt{Var}[x]. 
\end{split}
\end{equation}
\end{lemma}
The variance of SSP blocks is smaller than that of \textit{ResBlock}. The variance adds to our argument that the SSP property is the reason for improved robustness;
for more detailed derivation and proof, see the supplement.

\noindent\textbf{Adaptive SSP Networks.}\quad
\begin{figure}[t]
    \centering
    \includegraphics[width=\textwidth]{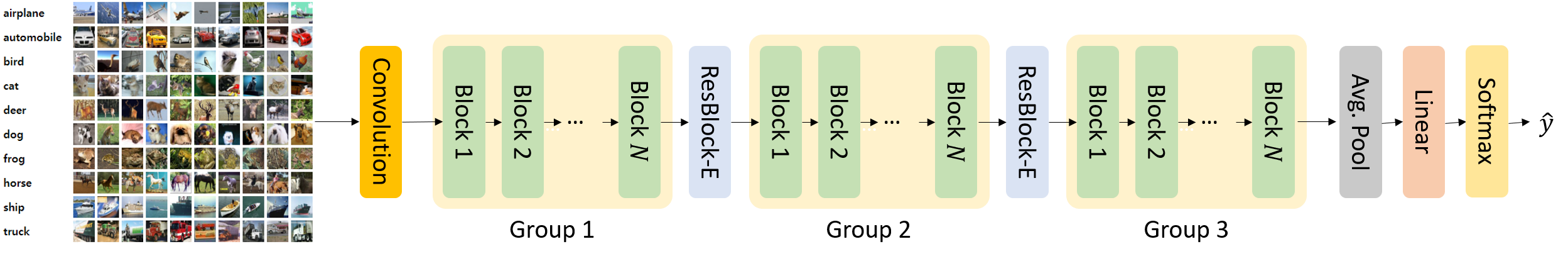}
    \caption{\myfontsize
    The overall architecture of neural networks used in experiments. Each group has $N \in \{6,10\}$ blocks and the block is either \textit{ResBlock}, SSP blocks (2 or 3) or \textit{ArkBlock}.
    The \textit{ResBlock-E} is inserted between groups to expand the number of channels for all the architectures.
    }
    \label{fig:model_figure}
\end{figure}
Also, we generalize Equation \eqref{eq:runge-kutta} with the second-order Adaptive Runge-Kutta block (\textit{ArkBlock}) that has the SSP property by construction.
These novel computational blocks slightly increase the number of parameters compared to \textit{ResBlock} but also provide greater robustness and natural accuracy than \textit{SSP2-Block} or \textit{SSP3-Block}.
Finally, we explore different computational architectures within each group to retain natural accuracy and further improve robustness.

A naive implementation of Equation \eqref{eq:runge-kutta} yields 5 additional parameters.
We can retain the SSP property in \textit{ArkBlocks} by reducing the number of parameters with Ralston's method \cite{gottlieb1998total}.
Thus, the number of additional learned parameters per block, when
compared with \textit{ResBlock},
is 2
and is defined as
\begin{equation}
    \label{eq:ralston}
    \begin{split}
        \alpha_{1,0} & = 1, \qquad \alpha_{2,0} =1-\alpha_{2,1},\\
        \beta_{2,0}&=1-\frac{1}{2\beta_{1,0}}-\alpha_{2,1}\beta_{1,0}, \qquad
        \beta_{2,1}=\frac{1}{2\beta_{1,0}}.
    \end{split}
\end{equation}
\textcolor{black}{We further improve performance by reducing the number of parameters by fixing $\alpha_{2,1}$ and simply learning $\beta_{1,0}$ in each block.}

Adaptive SSP networks still maintain the same architecture, as in Figure \ref{fig:model_figure}, but are comprised of blocks that have the form
\begin{equation}
    \label{eq:arkmodel}
    \begin{split}
        u^{(1)}&= u^n + \beta_{1,0}L(u^n), \\
        u^{(n+1)}&=\alpha_{2,0}u^n+\beta_{2,0}L(u^n)+\alpha_{2,1}u^{(1)}+\beta_{2,1}L(u^{(1)}).
    \end{split}
\end{equation}
We implement \textit{ArkBlock}s with,
\begin{equation}
\label{eq:ark_implement}
\begin{split}
    x({t_{k+\frac{1}{2}}}) &= x(t_k) + \beta_{1,0}F\left(x(t_k);\Theta(t_k)\right) ,    \\
    x(t_{k+1}) &= \alpha_{2,0}x(t_k) + \beta_{2,0} F\left(x({t_{k}});\Theta(t_k)\right)\\
    &~+ \alpha_{2,1} x({t_{k+\frac{1}{2}}}) + \beta_{2,1} F\left(x({t_{k+\frac{1}{2}}});\Theta(t_k)\right).
\end{split}
\end{equation}
The {\it ArkBlock}s are inspired by the generalized Runge-Kutta method in \eqref{eq:arkmodel}. 
However, the numerical scheme in Equation \eqref{eq:arkmodel}, keeps $\alpha_{2,1}$ and $\beta_{1,0}$ constant in all blocks, while {\it ArkBlock}s set those parameters as \textit{learnable}; varying in each block.
Such an adaptivity based on data and architectures cannot be obtained by mathematically derived coefficients.
To our knowledge, this is the first attempt.

\begin{table}[h]
\centering
\resizebox{0.5\columnwidth}{!}{
\begin{tabular}{c|c|c|c|c}
\toprule
Model & Clean & FGSM & PGD$_{20}$ & PGD$_{30}$ \\
\midrule\midrule
ResNet & 0.9961 & 0.7674 & 0.5799 & 0.1773 \\
SSP-2 & 0.9954 & 0.7984 & 0.5979 & 0.1850 \\ 
SSP-3 & 0.9960 & 0.8022 & 0.6176 & 0.1930 \\
SSP-adap & 0.9946 & \textbf{0.8586} & \textbf{0.7611} & \textbf{0.5102} \\

\bottomrule
\end{tabular}
}
\caption{\myfontsize The accuracy against adversarial attacks with standard training on the MNIST dataset; all models were trained with 6 blocks. Note that PGD$_{i}$ represents a projected gradient descent attack with $i$ iterations and that all the SSPNets are more robust against adversarial attacks than ResNet.
}
\label{tab:result_mnist}
\end{table}
\section{Experiments}
We evaluate the robustness of various SSP networks against adversarial examples.
MNIST~\cite{lecun-mnisthandwrittendigit-2010} and CIFAR10~\cite{krizhevsky2009learning} are used for evaluation; for results on other datasets, see the supplement.
The robustness is measured by the classification accuracy on adversarial examples generated by FGSM \cite{goodfellow2014explaining} and PGD \cite{madry2017towards}.

In this section, we empirically address the following three questions:
\begin{itemize}
    \item Are deep neural networks with the SSP property more robust than ResNet when the models are trained with or without adversarial training?
    \item Can we further improve upon adversarial robustness and simultaneously retain the natural accuracy of ResNet?
    \item Do Strong Stability Preserving networks suppress the perturbation growth during forward propagation? 
\end{itemize}

\subsection{Experimental setup}
\noindent\textbf{ResNet and SSP networks.}\quad
Each group has $N$ blocks where each block can be either \textit{ResBlock}, \textit{SSP2-block}, \textit{SSP3-block}, or \textit{ArkBlock}, as seen in Figure \ref{fig:model_figure}.
Networks are named after the type of blocks: ResNet, SSP-2, SSP-3, and SSP-adap.
The blocks in each group have the same number of input/output channels.
The convolutional layers in group 1, group 2, and group 3 have 16, 32, 64 channels respectively.
The classification layer of our networks consist of an average pooling and softmax layer, in order to calculate the confidence score.

\subsection{Evaluation on MNIST with standard training}
We demonstrate that SSPNets are more robust than ResNet with standard training. Since MNIST has relatively low-resolution images compared to CIFAR10, we used a smaller architecture by skipping group 1 and 2 in Figure \ref{fig:model_figure}.

\noindent\textbf{Experimental Details.}\quad
We evaluate the models on MNIST.
When training the models, samples are augmented by adding random noise $\delta$ drawn from a uniform distribution $Uniform(-\epsilon,\epsilon)$. We set the maximum perturbation magnitude $\epsilon=0.3$ for both training and evaluation. For optimization, Adam \cite{kingma2014adam} is used with learning rate $0.0001$ and $(\beta_1,\beta_2)=(0.9,0.999)$, minibatch size of 128. Models are trained for 100 epochs.

\noindent\textbf{Robustness Comparison.}\quad
The results in Table \ref{tab:result_mnist} show that 
all four models have high accuracy ($99.5\sim99.6\%$) in classifying clean samples.
This means that SSP blocks do not lead to a significant loss of accuracy on clean samples.
Further, the improvement by SSP compared to ResNet is consistently observed in different settings.
SSP-2 improves the robustness by 3\% against FGSM and 1\% against PGD. 
SSP-3 shows larger improvement about 4\% and 2\% against FGSM and PGD.
\textcolor{black}{SSP-adap shows the largest improvement about 9\% and 33\% against FGSM and PGD.}
It is known that adversarial training on MNIST is sufficiently robust against FGSM and PGD. 
All models trained by adversarial training achieve $96\sim97\%$ on MNIST, which makes it hard to demonstrate the benefit of SSP networks with adversarial training compared to ResNet. 

\subsection{SSP with adversarial training}
We analyze the robustness of SSP networks, on the CIFAR10 dataset.
Our preliminary experiments show that all the models, e.g., ResNet, SSP-2, SSP-3, and SSP-adap trained without adversarial training are easily fooled by 
PGD attacks, but more analysis is needed on a more challenging dataset.
For this reason, we focus on the adversarial training setting for CIFAR10. Please see supplementary materials for more analysis on SSP networks with adversarial training.
\begin{table*}[h]
    \centering
    \resizebox{0.6\columnwidth}{!}{
    \begin{tabular}{c|c|c|c|c|c|c|c}
        \toprule
        $N$    & $K$    & Model & Clean   & FGSM  & PGD$_{7}$ & PGD$_{12}$  & PGD$_{20}$  \\
        \midrule\midrule
        6 & 7   & ResNet    & 0.8357    & 0.5116    & 0.4389   & 0.4215 & 0.4150    \\
        6 & 7   & mid-RK2   & \textbf{0.8407}    & 0.5156    & 0.4377    & 0.4193    & 0.4129    \\
        6 & 7   & SSP-2    & 0.8257    & 0.5223    & 0.4577 & 0.4426 & 0.4368    \\
        6 & 7   & SSP-3    & 0.8376    & 0.5165    & 0.4478    & 0.4305 & 0.4246    \\
        6 & 7   & SSP-adap      & 0.8376    & \textbf{0.5283}    & \textbf{0.4640}    & \textbf{0.4455}  &  \textbf{0.4403}    \\
        \midrule
        6 & 12   & ResNet   & \textbf{0.8010}    & 0.5304    & 0.4817   & 0.4691    & 0.4650    \\
        6 & 12   & mid-RK2  & 0.7957    & 0.5326    & 0.4849    & 0.4740   & 0.4693    \\
        6 & 12   & SSP-2    & 0.7899    & 0.5426    & 0.5073    & 0.4983   & 0.4961    \\
        6 & 12   & SSP-3    & 0.7966    & 0.5440    & \textbf{0.5092}    & \textbf{0.4999}   & \textbf{0.4976}    \\
        6 & 12   & SSP-adap      & 0.7988    & \textbf{0.5504}    & 0.5066    & 0.4964   & 0.4943    \\
        \midrule
        10 & 7   & ResNet   & \textbf{0.8516}   & 0.5225    & 0.4398   & 0.4188  & 0.4111    \\
        10 & 7   & mid-RK2  & 0.8451    & 0.5146    & 0.4343   & 0.4122   & 0.4045    \\
        10 & 7   & SSP-2    & 0.8437    & \textbf{0.5373}    & 0.4714   & 0.4502   & 0.4427    \\
        10 & 7   & SSP-3    & 0.8505    & 0.5350    & \textbf{0.4719}   & \textbf{0.4558}   & \textbf{0.4497}    \\
        10 & 7   & SSP-adap      & 0.8504     & 0.5308    & 0.4592      & 0.4376   & 0.4310    \\
        \midrule
        10 & 12   & ResNet   & 0.8181    & 0.5467    & 0.4957   & 0.4799   & 0.4755    \\
        10 & 12   & mid-RK2  & \textbf{0.8198}    & 0.5522    & 0.4968   & 0.4818   & 0.4775    \\
        10 & 12   & SSP-2    & 0.8144    & 0.5497    & 0.5074   & 0.4957   & 0.4932    \\
        10 & 12   & SSP-3    & 0.8119    & 0.5507    & 0.5032   & 0.4929   & 0.4890    \\
        10 & 12   & SSP-adap      & 0.8156    & \textbf{0.5643}    & \textbf{0.5166}    & \textbf{0.5054}  & \textbf{0.5016}    \\
        \bottomrule
    \end{tabular}
    }
    \caption{\myfontsize CIFAR10 robustness evaluation against adversarial attacks. The column index $N$ indicates the number of blocks in each group of Figure \ref{fig:model_figure}, $K$ indicates the number of PGD iterations during training while PGD$_i$ represents the attack with $i$ iterations during attack.
    The SSP-adap model indicates an adaptive Runge-Kutta structure.
    All the SSP networks are more robust against adversarial attack than ResNet.
    Moreover, SSP-adap maintains the natural accuracy.
    }
    \label{tab:result}
    \vspace{-20pt}
\end{table*}

\noindent\textbf{Adversarial Training.}\quad
Before experimental results, we briefly summarize the adversarial training proposed by \cite{madry2017towards}.
The objective of adversarial training is to minimize the adversarial risk given as,
\begin{equation}
\label{eq:adv_risk}
    R_{adv}(h_\theta) = \mathbb{E}_{(x,y)\sim D}\big[\max\limits_{\delta \in \Delta} \mathcal{L}(h_\theta (x+\delta),y)\big],
\end{equation}
where the $h_{\theta}$ is a model parameterized by $\theta$, $\mathcal{L}$ is a loss function, $y$ is the label of corresponding image $x$, $D$ is a true data distribution, and $\Delta$ is a set of small perturbations satisfying $\|\delta\|_{p}\le\epsilon$.
In our experiments, the $\ell_{\infty}$ metric is used, i.e., $p=\infty$.
Finding the exact solution to $\max\limits_{\delta \in \Delta} \mathcal{L}(h_\theta (x+\delta),y)$ is intractable, so \cite{madry2017towards} approximate it with a sample generated by the PGD attack.
PGD attack finds the adversarial example given as 
$x_{i+1} = \Pi(x_i + \alpha \nabla_{x_i} \mathcal{L}(h_{\theta}(x_i),y))$,
where $i\in\{0,1,\cdots,K-1\}$, $K$ is the number of iterations of PGD attack, $\Pi$ denotes the projection to a small ball $\Delta$ and a valid pixel range.
In our experiment, $x_0$ is initialized with the input image augmented by adding the random perturbation $\delta_0$ sampled from the uniform distribution $Uniform(-\epsilon,\epsilon)$.

To summarize, our adversarial training procedure works as follows:
First, randomly perturb the image within the allowed perturbation range $\epsilon$.
Next, generate the candidate adversarial example by PGD attack. 
Finally, take the gradient descent step on a minibatch composed of only candidate adversarial examples.
The adversarial training is closely related to the Frank-Wolfe Algorithm and two projections in the original adversarial training can be simplified to one projection to the intersection of two convex sets.
The pseudocode of adversarial training and a detailed discussion of implementation are provided in the supplement.

\noindent\textbf{Experimental Details.}\quad
We use the Stochastic Gradient Descent method with Nesterov momentum, learning rate of $0.1$, weight decay of $0.0005$, momentum $0.9$, and a minibatch size of 128 samples.
All models are trained for 200 epochs and in every 60, 100, 140 epochs, the learning rate decayed with a decaying factor $0.1$.
Both adversarial training and robustness evaluation, we set the maximum perturbation range $\epsilon=8/255$.
To evaluate the robustness, we use FGSM \cite{goodfellow2014explaining} and PGD \cite{madry2017towards}; similar to our MNIST experiments.
We set the PGD attack parameters to $\alpha=2/255$, and the number of iterations $K=7,12,20$ in evaluation.


\noindent\textbf{Robustness Comparison.}\quad The experimental results are shown in Table \ref{tab:result}.
Models are evaluated in four different settings varying both the number of blocks (6 or 10 in column $N$) for each group in Figure \ref{fig:model_figure} and the number of iterations in PGD (7 or 12 in column $K$) to generate adversarial examples during training. 

Before discussing about the effectiveness of SSP, we briefly show the relationship among robustness, the amount of model parameters, and the strength of attacks used in adversarial training. 
As shown in Table \ref{tab:result}, the robustness of all the models is improved by stronger attacks during training  (e.g., larger $K$ in PGD). The same observation is reported in  \cite{wang2019on}. For instance, SSP-3 (N=6, K=12) shows higher accuracy than SSP-3 (N=6, K=7) against all the attacks and especially the improvement is about 7\% against PGD with $20$ iterations.
Also, a bigger model size (e.g., larger $N$) increases the robustness against adversarial examples.
This is closely related to the finding in \cite{madry2017towards} that increasing the number of channels in hidden layers often improves the robustness.
Our experiments show that increasing the model size by adding more layers improves the robustness. For example, when $K=7$, SSP-3 with $N=10$ blocks show overall higher accuracy than SSP-3 with $N=6$ blocks, the gain is about $2 \%$. From the numerical discretization perspective,  more blocks can be seen as a finer time discretization that leads to a more accurate numerical solution (or prediction).

All SSPNets, SSP-2, SSP-3 and SSP-adap, consistently outperform ResNet by, roughly, $1 \sim 3.9\%$ when ResNet and SSPNets have the same number of blocks, $N$, and iterations, $K$, in adversarial training.
Note that we compare SSP-2 (and SSP-3) with ResNet, which has the same amount of parameters, and this is important to assure that the gain is not from an increased the amount of model parameters.
Also, SSP-2, SSP-3 and SSP-adap have the same time discretization as ResNet.
So, we conclude that the improvement in robustness against adversarial attacks solely comes from the strength of a higher-order numerical discretization.
Table \ref{tab:result} shows one more interesting property of SSP networks.  Unlike adversarial training and defensive methods that usually cause the label leaking effects \cite{tsipras2018robustness,trades2019}, SSP-2, SSP-3 and SSP-adap (our architectural changes) do not bring any additional loss of accuracy on natural samples.

On the other hand, Table \ref{tab:result} also shows that the mid-RK2 architecture does not outperform ResNet, SSP-2, SSP-3 or SSP-adap even though the mid-RK2 is derived from the second order numerical scheme.
This gives credence to the implementation of SSPNets and implies that the robust performance is not a result of arbitrary high-order methods.
\textcolor{black}{
In addition, SSP-adap achieves comparable natural accuracy as ResNet and improves robustness. Table \ref{tab:result} demonstrates the consistent improvement across various settings. For example, SSP-adap achieves nearly 4\% absolute performance improvement for $N=10,K=7$.}
\textcolor{black}{The improvement by SSP networks compared to ResNet and the performance difference between different SSP networks are relatively smaller than Table \ref{tab:result_mnist}.
Our conjecture is that this is due to the improvement by adversarial training.}
We believe that the Strong Stability Preserving property imposed by our architectural change allows the SSPNets to improve the robustness against adversarial attacks.

\captionsetup[sub]{font=small,labelfont={bf,sf}}
\begin{figure}[t]
    \centering
    \begin{subfigure}{0.248\textwidth}
        \centering
         \includegraphics[width=\textwidth]{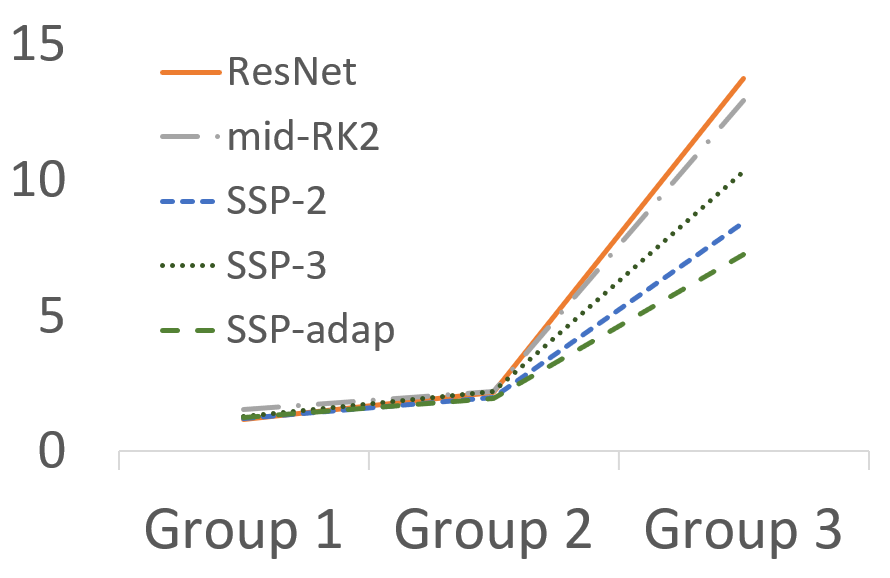}
        \caption{N$=$6,K$=$7,p$=$1}
    \end{subfigure}%
    \begin{subfigure}{0.248\textwidth}
        \centering
        \includegraphics[width=\textwidth]{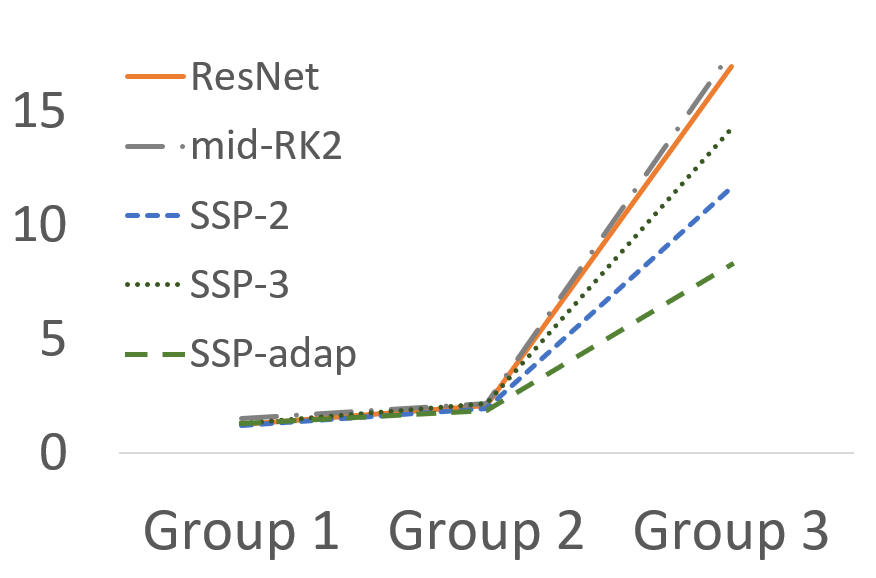}
        \caption{N$=$6,K$=$7,p$=$2}
    \end{subfigure}%
    \begin{subfigure}{0.248\textwidth}
        \centering
        \includegraphics[width=\textwidth]{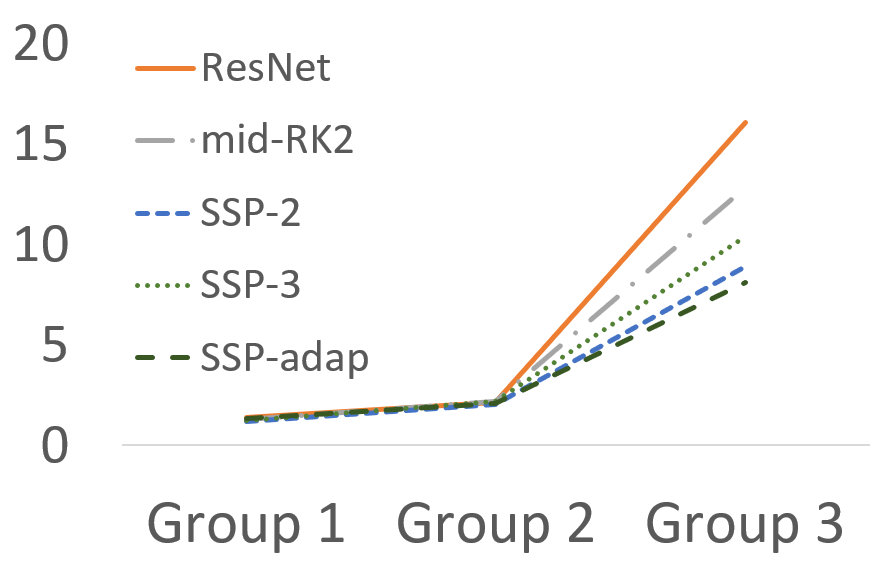}
        \caption{N$=$10,K$=$12,p$=$1}
    \end{subfigure}%
    \begin{subfigure}{0.248\textwidth}
        \centering
        \includegraphics[width=\textwidth]{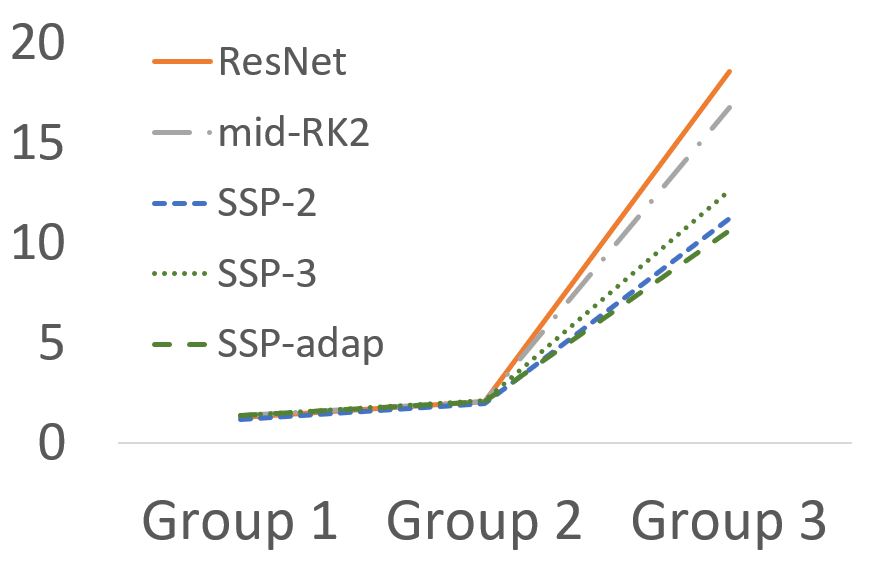}
        \caption{N$=$10,K$=$12,p$=$2}
    \end{subfigure}%
    \caption{\myfontsize 
    Perturbation growth ratio in Equation \eqref{eq:pgr} of clean samples and its adversarial counterparts. As the SSP networks suppress the perturbation growth during forward propagation, SSP-2, SSP-3 and SSP-adap have a lower ratio than ResNet. For full version of this figure, see the supplement.
    }
    \label{fig:normratio}
\end{figure}
\noindent\textbf{Perturbation Growth Ratio Comparison.}\quad
We investigate how the distance between clean samples and adversarial examples evolves through networks by calculating the perturbation growth ratio between input/output of groups given by
\begin{equation}
\label{eq:pgr}
    \texttt{PGR}(f)=\mathbb{E}_{x\sim\mathcal{D}}\left[\mathbb{E}_{x'\sim\mathcal{X'}}\left[\frac{\|f(x) - f(x')\|_{p}}{\|x - x'\|_{p}}\right]\right],\quad p\in\{1,2\}
\end{equation}
where $f(\cdot)$ is a function of a group, $x'$ is a corrupted sample from $x$ and is an element of the set $\mathcal{X'}$, $\mathcal{X'}$ is a small neighborhood of $x$, and \textcolor{black}{$p$ defines a type of norm either $\ell_1$ (related to TV in Equation \eqref{eq:TV}) or $\ell_2$ (related to Lemma \ref{lem:var}).}
Since each model has a different scale of feature maps, to compare, the distance needs proper normalization.
So, we first measure the distance between a clean sample and its adversarial example before/after each group in Figure \ref{fig:model_figure}.
$x'$ is the adversarial example generated by PGD attack with 20 iterations for each model.

Figure \ref{fig:normratio} presents the perturbation growth ratio when $N=6, K=7$ and $N=10, K=12$ at each group in the models.
Since the adversarial examples change the final predictions, the perturbation growth ratio increases in all the models. 
However, \textcolor{black}{for SSPNets, the perturbation growth ratio is significantly lower than ResNet.}
This result supports that the proposed SSP blocks improve robustness of networks against adversarial attacks when compared to ResNet.
We also conducted an experiment when $x'$ is corrupted by adding a random perturbation to $x$ and the result is consistent with Figure \ref{fig:normratio}.
For full version of Figure \ref{fig:normratio} and more discussion, see the supplement.


\section{Conclusion}
In this work, we leverage the Strong Stability Preserving property of numerical discretization in order to improve adversarial robustness.
Inspired by the Strong Stability Preserving methods,
we design a series of SSPNets by applying the same \textit{ResBlock} multiple times with parameters derived from numerical analysis.
All of the SSP networks provide robustness against adversarial attacks.
In particular, SSPNets with the \textit{ArkBlock} improve adversarial robustness while maintaining natural accuracy.
The proposed networks are complementary with adversarial training and suppress the perturbation growth.
Our work shows the way to improve the robustness of neural networks by utilizing the theory of advanced numerical discretization schemes.
We believe that the intersection of numerical discretization and robust deep learning will provide new opportunities to study robust neural networks.
\footnote[1]{The codes are available at \url{https://github.com/matbambbang/sspnet}.}

\footnotesize{
\noindent\textbf{Acknowledgements}\quad
This work was supported by Institute of Information \& communications Technology Planning \& Evaluation (IITP) grant funded by the Korea government (MSIT) (No.2019-0-00533, Research on CPU vulnerability detection and validation), National Supercomputing Center with supercomputing resources including technical support (KSC-2019-CRE-0186), National Research Foundation of Korea (NRF-2020R1A2C3010638), and Simons Foundation Collaboration Grants for Mathematicians.
}

\bibliographystyle{splncs04}
\bibliography{reference}

\begin{thebibliography}{10}
\providecommand{\url}[1]{\texttt{#1}}
\providecommand{\urlprefix}{URL }
\providecommand{\doi}[1]{https://doi.org/#1}

\bibitem{athalye2018obfuscated}
Athalye, A., Carlini, N., Wagner, D.: Obfuscated gradients give a false sense
  of security: Circumventing defenses to adversarial examples. In: ICML. pp.
  274--283 (2018)

\bibitem{ben2009robust}
Ben-Tal, A., El~Ghaoui, L., Nemirovski, A.: Robust optimization, vol.~28.
  Princeton University Press (2009)

\bibitem{buckman2018thermometer}
Buckman, J., Roy, A., Raffel, C., Goodfellow, I.: Thermometer encoding: One hot
  way to resist adversarial examples. In: ICLR (2018)

\bibitem{Butcher87textbook}
Butcher, J.C.: The numerical analysis of ordinary differential equations. A
  Wiley-Interscience Publication, John Wiley \& Sons, Ltd., Chichester (1987),
  runge Kutta and general linear methods

\bibitem{carlini2017towards}
Carlini, N., Wagner, D.: Towards evaluating the robustness of neural networks.
  In: 2017 IEEE Symposium on Security and Privacy (SP). pp. 39--57 (2017)

\bibitem{chen2018neural}
Chen, T.Q., Rubanova, Y., Bettencourt, J., Duvenaud, D.K.: Neural ordinary
  differential equations. In: NeurIPS. pp. 6572--6583 (2018)

\bibitem{ciccone2018nais}
Ciccone, M., Gallieri, M., Masci, J., Osendorfer, C., Gomez, F.: {NAIS-N}et:
  stable deep networks from non-autonomous differential equations. In: NeurIPS.
  pp. 3025--3035 (2018)

\bibitem{goodfellow2014explaining}
Goodfellow, I., Shlens, J., Szegedy, C.: Explaining and harnessing adversarial
  examples. In: ICLR (2015)

\bibitem{gottlieb1998total}
Gottlieb, S., Shu, C.W.: Total variation diminishing runge-kutta schemes.
  Mathematics of computation of the American Mathematical Society
  \textbf{67}(221),  73--85 (1998)

\bibitem{gottlieb2001Siam}
Gottlieb, S., Shu, C.W., Tadmor, E.: Strong stability-preserving high-order
  time discretization methods. SIAM Rev.  \textbf{43}(1),  89--112 (2001)

\bibitem{grathwohl2018ffjord}
Grathwohl, W., Chen, R.T., Betterncourt, J., Sutskever, I., Duvenaud, D.:
  {FFJORD}: Free-form continuous dynamics for scalable reversible generative
  models. arXiv preprint arXiv:1810.01367  (2018)

\bibitem{Harten83}
Harten, A.: High resolution schemes for hyperbolic conservation laws. J.
  Comput. Phys.  \textbf{49}(3),  357--393 (1983)

\bibitem{Harten87}
Harten, A., Engquist, B., Osher, S., Chakravarthy, S.R.: Uniformly high-order
  accurate essentially nonoscillatory schemes. {III}. J. Comput. Phys.
  \textbf{71}(2),  231--303 (1987)

\bibitem{he2015delving}
He, K., Zhang, X., Ren, S., Sun, J.: Delving deep into rectifiers: Surpassing
  human-level performance on imagenet classification. In: Proceedings of the
  IEEE international conference on computer vision. pp. 1026--1034 (2015)

\bibitem{he2016deep}
He, K., Zhang, X., Ren, S., Sun, J.: Deep residual learning for image
  recognition. In: CVPR. pp. 770--778 (2016)

\bibitem{he2016identity}
He, K., Zhang, X., Ren, S., Sun, J.: Identity mappings in deep residual
  networks. In: ECCV. pp. 630--645. Springer (2016)

\bibitem{kingma2014adam}
Kingma, D.P., Ba, J.: Adam: A method for stochastic optimization. In: ICLR
  (2014)

\bibitem{krizhevsky2009learning}
Krizhevsky, A., Hinton, G.: Learning multiple layers of features from tiny
  images. Tech. rep., Citeseer (2009)

\bibitem{lecun-mnisthandwrittendigit-2010}
LeCun, Y., Cortes, C.: {MNIST} handwritten digit database  (2010),
  \url{http://yann.lecun.com/exdb/mnist/}

\bibitem{lu2017beyond}
Lu, Y., Zhong, A., Li, Q., Dong, B.: Beyond finite layer neural networks:
  Bridging deep architectures and numerical differential equations. In: ICML.
  pp. 5181--5190 (2018)

\bibitem{madry2017towards}
Madry, A., Makelov, A., Schmidt, L., Tsipras, D., Vladu, A.: Towards deep
  learning models resistant to adversarial attacks. In: ICLR (2018)

\bibitem{papernot2016distillation}
Papernot, N., McDaniel, P., Wu, X., Jha, S., Swami, A.: Distillation as a
  defense to adversarial perturbations against deep neural networks. In: 2016
  IEEE Symposium on Security and Privacy (SP). pp. 582--597. IEEE (2016)

\bibitem{raff2019bart}
Raff, E., Sylvester, J., Forsyth, S., McLean, M.: Barrage of random transforms
  for adversarially robust defense. In: CVPR (June 2019)

\bibitem{rubanova2019latent}
Rubanova, Y., Chen, R.T., Duvenaud, D.: Latent odes for irregularly-sampled
  time series. arXiv preprint arXiv:1907.03907  (2019)

\bibitem{ruthotto2018deep}
Ruthotto, L., Haber, E.: Deep neural networks motivated by partial differential
  equations. arXiv preprint arXiv:1804.04272  (2018)

\bibitem{samangouei2018defensegan}
Samangouei, P., Kabkab, M., Chellappa, R.: Defense-{GAN}: Protecting
  classifiers against adversarial attacks using generative models. In: ICLR
  (2018)

\bibitem{shu1988total}
Shu, C.W.: Total-variation-diminishing time discretizations. SIAM Journal on
  Scientific and Statistical Computing  \textbf{9}(6),  1073--1084 (1988)

\bibitem{shu1988efficient}
Shu, C.W., Osher, S.: Efficient implementation of essentially non-oscillatory
  shock-capturing schemes. Journal of computational physics  \textbf{77}(2),
  439--471 (1988)

\bibitem{song2018pixeldefend}
Song, Y., Kim, T., Nowozin, S., Ermon, S., Kushman, N.: Pixeldefend: Leveraging
  generative models to understand and defend against adversarial examples. In:
  ICLR (2018)

\bibitem{szegedy2013intriguing}
Szegedy, C., Zaremba, W., Sutskever, I., Bruna, J., Erhan, D., Goodfellow, I.,
  Fergus, R.: Intriguing properties of neural networks. arXiv preprint
  arXiv:1312.6199  (2013)

\bibitem{tsipras2018robustness}
Tsipras, D., Santurkar, S., Engstrom, L., Turner, A., Madry, A.: Robustness may
  be at odds with accuracy. In: ICLR (2019)

\bibitem{wang2019on}
Wang, Y., Ma, X., Bailey, J., Yi, J., Zhou, B., Gu, Q.: On the convergence and
  robustness of adversarial training. In: ICML. pp. 6586--6595 (2019)

\bibitem{wong2017provable}
Wong, E., Kolter, Z.: Provable defenses against adversarial examples via the
  convex outer adversarial polytope. In: ICML. pp. 5283--5292 (2018)

\bibitem{wong2018scaling}
Wong, E., Schmidt, F., Metzen, J.H., Kolter, J.Z.: Scaling provable adversarial
  defenses. In: NeurIPS. pp. 8400--8409 (2018)

\bibitem{xie2019feature}
Xie, C., Wu, Y., Maaten, L.v.d., Yuille, A.L., He, K.: Feature denoising for
  improving adversarial robustness. In: CVPR. pp. 501--509 (2019)

\bibitem{xu2009robustness}
Xu, H., Caramanis, C., Mannor, S.: Robustness and regularization of support
  vector machines. Journal of Machine Learning Research  \textbf{10}(Jul),
  1485--1510 (2009)

\bibitem{yang2019me}
Yang, Y., Zhang, G., Xu, Z., Katabi, D.: Me-net: Towards effective adversarial
  robustness with matrix estimation. In: ICML. pp. 7025--7034 (2019)

\bibitem{trades2019}
Zhang, H., Yu, Y., Jiao, J., Xing, E., Ghaoui, L.E., Jordan, M.: Theoretically
  principled trade-off between robustness and accuracy. In: ICML. pp.
  7472--7482 (2019)

\bibitem{zhang2018residual}
Zhang, H., Dauphin, Y.N., Ma, T.: Residual learning without normalization via
  better initialization. In: International Conference on Learning
  Representations (2019)

\bibitem{zhang2017polynet}
Zhang, X., Li, Z., Change~Loy, C., Lin, D.: Polynet: A pursuit of structural
  diversity in very deep networks. In: CVPR. pp. 718--726 (2017)

\end{thebibliography}

\appendix

\section{Summary}
This supplementary material is structured as follows: proofs of strong stability preserving methods (section~\ref{sec:proof}),
proofs of variance analysis of SSP networks (section~\ref{sec:var}),
comparison of non-TVD scheme and TVD scheme (section~\ref{sec:tvdcompare}),
reminder of adversarial training (section~\ref{sec:adversarial_training}),
exploratory network analysis (section~\ref{sec:netanalysis})
and suppression on perturbation growth (section~\ref{sec:suppression}).
\section{Proofs of Strong Stability Preserving Scheme}
\label{sec:proof}

\begin{lemma}
\cite{shu1988efficient} If the forward Euler method is strongly stable under the CFL condition, i.e. $||u^n+\Delta tL(u^n)|| \le ||u^n||$, then the Runge-Kutta method possesses SSP, $||u^{n+1}|| \le ||u^n||$, provided that $\Delta t \le c\Delta t_{CFL}.$
\end{lemma}
{\bf Sketch of proof.}
To begin, we rewrite the Runge-Kutta method as a convex combination of forward Euler steps
\begin{equation*}
\begin{split}
    \| u^{(i)} \| 
    & = \left\| \sum^{i-1}_{k=0} \left( \alpha_{i,k} u^{(k)} + \Delta t \beta_{i,k} L(u^{(k)}) \right)  \right \|\\
    & \leq \sum^{i-1}_{k=0} \alpha_{i,j} \left\| u^{(k)} + \Delta t \frac{\beta_{i,k}}{\alpha_{i,k}} L(u^{(k)}) \right\|.
\end{split}    
\end{equation*}
If we set $c = \min_{i,k} (\alpha_{i,k}/\beta_{i,k})$ for $\Delta t \le c\Delta t_{CFL}$, we find that
\begin{equation*}
    \| u^{(k)} + \Delta t \frac{\beta_{i,k}}{\alpha_{i,k}} L(u^{(k)}) \| \leq \| u^{(k)} \|.
\end{equation*}
Also, we notice that $\sum^{i-1}_{k=0} \alpha_{i,k} = 1$ by consistency.
We now use induction to show
\begin{equation} \label{e:app1}
    \| u^{(k)} \| \leq \| u^n \|,
\end{equation}
for $k=0,1,...,m$. 
Clearly, when $k=0$, \eqref{e:app1} holds. Assuming that it is valid for all $k \leq i-1$, we deduce that
\begin{equation*}
\begin{split}
    \| u^{(i)} \| 
    & \leq 
    \sum^{i-1}_{k=0} \alpha_{i,j} \left\| u^{(k)} + \Delta t \frac{\beta_{i,k}}{\alpha_{i,k}} L(u^{(k)}) \right\| \\
    & \leq
    \sum^{i-1}_{k=0} \alpha_{i,k} \| u^{(k)} \| \\
    & \leq 
     \sum^{i-1}_{k=0} \alpha_{i,k} \| u^n \| = \| u^n \|.
\end{split}    
\end{equation*}
Hence, the lemma follows.

\begin{lemma}
\cite{shu1988efficient} An optimal second-order SSP Runge-Kutta method is given by,
\begin{equation}
\label{eq:ssp2}
\begin{split}
    & u^{(1)} = u^n + \Delta t L(u^n),    \\
    & u^{n+1} = \frac{1}{2}u^n + \frac{1}{2}u^{(1)}+\frac{1}{2}\Delta tL(u^{(1)}),
\end{split}
\end{equation}
with a CFL coefficient $c=1$.
In addition, an optimal third-order SSP Runge-Kutta method is of the form
\begin{equation}
\label{eq:ssp3}
\begin{split}
    & u^{(1)} = u^n + \Delta tL(u^n),    \\
    & u^{(2)} = \frac{3}{4}u^n + \frac{1}{4}u^{(1)}+\frac{1}{4}\Delta tL(u^{(1)}),   \\
    & u^{n+1} = \frac{1}{3}u^n + \frac{2}{3} u^{(2)} + \frac{2}{3} \Delta tL(u^{(2)}),
\end{split}
\end{equation}
with a CFL coefficient $c=1$.
\end{lemma}
{\bf Sketch of proof.}
For the second order $m=2$, we choose the coefficients as
\begin{equation*}
\begin{cases}
    \alpha_{1,0} = 1,\\
    \alpha_{2,0} = 1 - \alpha_{2,1},\\
    \beta_{2,0} = 1 - \frac{1}{2 \beta_{1,0}} - \alpha_{2,1} \beta_{1,0},\\
    \beta_{2,1} = \frac{1}{2 \beta_{1,0}},
\end{cases}    
\end{equation*}
where $\beta_{1,0}$ and $\alpha_{2,1}$ are free parameters.
Assume a CFL coefficient $c>1$, then $\alpha_{1,0} = 1$ implies $\beta_{1,0} < 1$.
Hence, we deduce that 
$$ \frac{1}{2 \beta_{1,0}}> \frac{1}{2}.$$
In addition, we note that
\begin{equation*}
    \alpha_{2,1} > \beta_{2,1}  = \frac{1}{2 \beta_{1,0}}
    \implies
    \alpha_{2,1} \beta_{1,0} > \frac{1}{2}.
\end{equation*}
Hence, we obtain that
\begin{equation*}
    \beta_{2,0} = 1 - \frac{1}{2 \beta_{1,0}} - \alpha_{2,1} \beta_{1,0} < 1 - \frac{1}{2} - \frac{1}{2} = 0,
\end{equation*}
which is a contradiction.
For the third order case $m=3$, we choose the coefficients as
\begin{equation*}
\begin{cases}
    \alpha_{3,2} = 1 - \alpha_{3,1} - \alpha_{3,0},\\
    \beta_{3,2} = \dfrac{3 \beta_{1,0}-2}{6 P (\beta_{1,0} - P)},\\
    \beta_{2,1} = \dfrac{1}{6 \beta_{1,0} \beta_{3,2}},\\
    \beta_{3,1} = \dfrac{1/2 - \alpha_{3,2} \beta_{1,0} \beta_{2,1} - P \beta_{3,2}}{\beta_{1,0}},\\
    \beta_{3,0} = 1 - \alpha_{3,1} \beta_{1,0} - \alpha_{3,2} P - \beta_{3,1} - \beta_{3,2},\\
    \beta_{2,0} = P - \alpha_{2,1} \beta_{1,0} - \beta_{2,1},
\end{cases}
\end{equation*}
where $\alpha_{2,1}, \alpha_{3,0}, \alpha_{3,1}, \beta_{1,0}$, and $P = \beta_{2,0} + \alpha_{2,1} \beta_{1,0} + \beta_{2,1}$ are free parameters.
We omit the detailed proof for the third order scheme as it is more technical. For the complete proof, see e.g. \cite{gottlieb1998total}.

\section{Proofs of Variance}
\label{sec:var}

In this section, we provide a proof of the Lemma \ref{lem:var}.


\begin{lemma}
\label{lem:var}
If $\texttt{Var}[F(x)] = \texttt{Var}[x]$, $\texttt{Cov}[x,F(y)] = 0$ then the variance increases by
\begin{equation}
\label{eq:variance}
\begin{split}
    & \texttt{Var}[ResBlock(x)] \;\;= 2\texttt{Var}[x], \;\;\;\;\;\; \texttt{Var}[\textit{mid-RK2(x)}] \;\;\;\:= \frac{9}{4}\texttt{Var}[x], \\
    & \texttt{Var}[\textit{SSP2-Block(x)}] = \frac{7}{4}\texttt{Var}[x], \;\;\;\;\; \texttt{Var}[\textit{SSP3-Block(x)}] = \frac{29}{18}\texttt{Var}[x]. 
\end{split}
\end{equation}
\end{lemma}

\noindent\textbf{Proof}\quad
To begin, we summarize basic properties of variance and convariance which commonly used in this proof.
\begin{equation}
\label{var_property}
\begin{split}
    & \texttt{Var}[x+y] = \texttt{Var}[x] + \texttt{Var}[y] + 2\texttt{Cov}[x,y], \\
    & \texttt{Var}[ax] = a^2\texttt{Var}[x],
\end{split}
\end{equation}

\begin{equation}
\label{cov_property}
\begin{split}
    & \texttt{Cov}[x,y+z] = \texttt{Cov}[x,y] + \texttt{Cov}[x,z], \\
    & \texttt{Cov}[ax,by] = ab\texttt{Cov}[x,y],
\end{split}
\end{equation}

where $a,b$ are real-valued constants, $x,y,z$ are random variables.

Our assumption holds
\begin{equation}
\label{assumption1}
    \texttt{Var}[F(x)] = \texttt{Var}[x]
\end{equation}
and
\begin{equation}
\label{assumption2}
    \texttt{Cov}[x,F(y;\Theta)] = 0,
\end{equation}
where $x$ and $y$ are random variables~\cite{he2015delving,zhang2018residual}.
Recall that operations of each block is written as

\begin{equation}
\label{eq:resblock}
    x_{k+1} = x_k + F(x_k;\Theta_k),
\end{equation}

\begin{equation}
\label{eq:midrk2}
\begin{split}
    & x_{k+\frac{1}{2}} = x_k + \frac{1}{2} F(x_k;\Theta_k), \\
    & x_{k+1} = x_k + F(x_{k+\frac{1}{2}};\Theta_k),
\end{split}
\end{equation}

\begin{equation}
\label{eq:sspblock2}
\begin{split}
    & x_{k+\frac{1}{2}}=x_k + F(x_k; \Theta_k), \\
    & x_{k+1} = \frac{1}{2}x_k + \frac{1}{2}x_{k+\frac{1}{2}}+\frac{1}{2}F(x_{k+\frac{1}{2}};\Theta_k),
\end{split}
\end{equation}

\begin{equation}
\label{eq:sspblock3}
\begin{split}
    & x_{k+\frac{1}{3}} = x_k + F(x_k;\Theta_k), \\
    & x_{k+\frac{2}{3}} = \frac{3}{4}x_k + \frac{1}{4}x_{k+\frac{1}{3}}+\frac{1}{4}F(x_{k+\frac{1}{3}};\Theta_k), \\
    & x_{k+1} = \frac{1}{3}x_k + \frac{2}{3}x_{k+\frac{2}{3}} + \frac{2}{3} F(x_{k+\frac{2}{3}};\Theta_k),
\end{split}
\end{equation}

where the Equation \eqref{eq:resblock} is the equation of \textit{ResBlock}, Equation \eqref{eq:midrk2} is \textit{mid-RK2 block}, Equation \eqref{eq:sspblock2} is \textit{SSP2-block}, Equation \eqref{eq:sspblock3} is \textit{SSP3-block}.
We divide the proofs of each block.
In every proofs, for simplicity, $F(x;\Theta_k) := f(x)$.

\noindent\textbf{Proof of \textit{ResBlock}}\quad
Using \eqref{assumption1} and \eqref{assumption2}, we derive the variance of output~\cite{zhang2018residual}.

\begin{equation*}
\begin{split}
    \texttt{Var}[x_{k+1}]
    & \equalref^{\eqref{var_property}} \texttt{Var}[x_k] + \texttt{Var}[f(x_k)] + 2\texttt{Cov}[x_k,f(x_k)] \\
    & \equalref^{\eqref{assumption2}} \texttt{Var}[x_k] + \texttt{Var}[f(x_k)] \\
    & \equalref^{\eqref{assumption1}} 2\texttt{Var}[x_k].
\end{split}
\end{equation*}

\noindent\textbf{Proof of \textit{mid-RK2}}\quad
First, we derive $\texttt{Var}[x_{k+\frac{1}{2}}]$ and $\texttt{Cov}[x_k,x_{k+\frac{1}{2}}]$.
\begin{equation}
\label{midrk2_1/2var}
\begin{split}
    \texttt{Var}[x_{k+\frac{1}{2}}]
    & \equalref^{\eqref{var_property}} \texttt{Var}[x_k] + \frac{1}{4}\texttt{Var}[f(x_k)] + \texttt{Cov}[x_k,f(x_k)] \\
    & \equalref^{(\ref{assumption1},\ref{assumption2})} \frac{5}{4}\texttt{Var}[x_k],
\end{split}
\end{equation}

\begin{equation}
\label{midrk2_1/2cov}
\begin{split}
    \texttt{Cov}[x_k,x_{k+\frac{1}{2}}] & = \texttt{Cov}[x_k, x_k + \frac{1}{2}f(x_k)] \\
    & \equalref^{\eqref{cov_property}} \texttt{Cov}[x_k,x_k] + \frac{1}{2}\texttt{Cov}[x_k, f(x_k)] \\
    & \equalref^{(\ref{assumption1},\ref{assumption2})} \texttt{Var}[x_k].
\end{split}
\end{equation}

By using $x_{k+1} = x_k + f(x_{k+\frac{1}{2}})$, 

\begin{equation*}
\begin{split}
    \texttt{Var}[x_{k+1}] & \equalref^{\eqref{eq:midrk2}} \texttt{Var}[x_k + f(x_{k+\frac{1}{2}})] \\
    & \equalref^{\eqref{var_property}} \texttt{Var}[x_k] + \texttt{Var}[f(x_{k+\frac{1}{2}})] + 2\texttt{Cov}[x_k, f(x_{k+\frac{1}{2}})] \\
    & \equalref^{(\ref{assumption1},\ref{assumption2})} \texttt{Var}[x_k] + \texttt{Var}[x_{k+\frac{1}{2}}] \\
    & \equalref^{\eqref{midrk2_1/2var}} \frac{9}{4}\texttt{Var}[x_k].
\end{split}
\end{equation*}
\noindent\textbf{Proof of \textit{SSP2-block}}\quad
We start with obtaining $\texttt{Var}[x_{k+\frac{1}{2}}]$ and $\texttt{Cov}[x_k,x_{k+\frac{1}{2}}]$.
\begin{equation}
\label{ssp2_1/2var}
\begin{split}
    \texttt{Var}[x_{k+\frac{1}{2}}]
    & \equalref^{\eqref{var_property}} \texttt{Var}[x_k] + \texttt{Var}[f(x_k)] + 2\texttt{Cov}[x_k,f(x_k)] \\
    & \equalref^{(\ref{assumption1},\ref{assumption2})} 2\texttt{Var}[x_k],
\end{split}
\end{equation}

\begin{equation}
\label{ssp2_1/2cov}
\begin{split}
    \texttt{Cov}[x_k,x_{k+\frac{1}{2}}]
    & \equalref^{\eqref{cov_property}} \texttt{Cov}[x_k,x_k] + \texttt{Cov}[x_k,f(x_k)] \\
    & \equalref^{\eqref{assumption2}} \texttt{Cov}[x_k,x_k] \\
    & = \texttt{Var}[x_k].
\end{split}
\end{equation}

Next, let $x^{(1)} = x_{k+\frac{1}{2}} + f(x_{k+\frac{1}{2}})$. Then we can derive $\texttt{Var}[x^{(1)}]$ and $\texttt{Cov}[x_k,x^{(1)}]$.


\begin{equation}
\label{ssp2_(1)var}
\begin{split}
    & \texttt{Var}[x^{(1)}] \\
    & \equalref^{\eqref{var_property}} \texttt{Var}[x_{k+\frac{1}{2}}] + \texttt{Var}[f(x_{k+\frac{1}{2}})] + 2\texttt{Cov}[x_{k+\frac{1}{2}},f(x_{k+\frac{1}{2}})] \\
    & \equalref^{\eqref{assumption2}} \texttt{Var}[x_{k+\frac{1}{2}}] + \texttt{Var}[f(x_{k+\frac{1}{2}})] \\
    & \equalref^{\eqref{assumption1}} \texttt{Var}[x_{k+\frac{1}{2}}] + \texttt{Var}[x_{k+\frac{1}{2}}] \\
    & = 2\texttt{Var}[x_{k+\frac{1}{2}}] \\
    & \equalref^{\eqref{ssp2_1/2var}} 4\texttt{Var}[x_k],
\end{split}
\end{equation}

\begin{equation}
\label{ssp2_(1)cov}
\begin{split}
    \texttt{Cov}[x_k,x^{(1)}]
    & \equalref^{\eqref{cov_property}} \texttt{Cov}[x_k,x_{k+\frac{1}{2}}] + \texttt{Cov}[x_k,f(x_{k+\frac{1}{2}})] \\
    & \equalref^{\eqref{assumption2}} \texttt{Cov}[x_k,x_{k+\frac{1}{2}}] \\
    & \equalref^{\eqref{ssp2_1/2cov}} \texttt{Var}[x_k].
\end{split}    
\end{equation}

Finally, by using $x^{(1)} = x_{k+\frac{1}{2}} + f(x_{k+\frac{1}{2}})$,
\begin{equation*}
\begin{split}
    \texttt{Var}[x_{k+1}]
    & \equalref^{\eqref{eq:sspblock2}} \texttt{Var}\left[\frac{1}{2}x_k + \frac{1}{2}x_{k+\frac{1}{2}} + \frac{1}{2}f(x_{k+\frac{1}{2}})\right] \\
    & = \texttt{Var}\left[\frac{1}{2}x_k + \frac{1}{2}x^{(1)}\right] \\
    & \equalref^{\eqref{var_property}} \frac{1}{4}\texttt{Var}[x_k] + \frac{1}{4}\texttt{Var}[x^{(1)}] + \frac{1}{2}\texttt{Cov}[x_k,x^{(1)}] \\
    & \equalref^{(\ref{ssp2_(1)var},\ref{ssp2_(1)cov})} \frac{1}{4}\texttt{Var}[x_k] + \texttt{Var}[x_k] + \frac{1}{2}\texttt{Var}[x_k] \\
    & = \frac{7}{4}\texttt{Var}[x_k].
\end{split}
\end{equation*}

\noindent\textbf{Proof of \textit{SSP3-block}}\quad
Similar to prove the \textit{SSP2-block}, the first step is inducing $\texttt{Var}[x_{k+\frac{1}{3}}]$ and $\texttt{Cov}[x_k,x_{k+\frac{1}{3}}]$.
\begin{equation}
\label{ssp3_1/3var}
    \texttt{Var}[x_{k+\frac{1}{3}}] = 2\texttt{Var}[x_k],
\end{equation}

\begin{equation}
\label{ssp3_1/3cov}
\begin{split}
    \texttt{Cov}[x_k,x_{k+\frac{1}{3}}]
    & \equalref^{\eqref{cov_property}} \texttt{Cov}[x_k,x_k] + \texttt{Cov}[x_k,f(x_k)] \\
    & \equalref^{\eqref{assumption1}} \texttt{Var}[x_k].
\end{split}
\end{equation}

Let $x^{(1)} = x_{k+\frac{1}{3}}+f(x_{k+\frac{1}{3}})$.
Then,
\begin{equation}
\label{ssp3_(1)var}
\begin{split}
    & \texttt{Var}[x^{(1)}] \\
    & \equalref^{\eqref{var_property}} \texttt{Var}[x_{k+\frac{1}{3}}] + \texttt{Var}[f(x_{x+\frac{1}{3}})] + 2\texttt{Cov}[x_{k+\frac{1}{3}},f(x_{x+\frac{1}{3}})] \\
    & \equalref^{\eqref{assumption2}} \texttt{Var}[x_{k+\frac{1}{3}}] + \texttt{Var}[f(x_{x+\frac{1}{3}})] \\
    & \equalref^{\eqref{assumption1}} 2\texttt{Var}[x_{k+\frac{1}{3}}] \\
    & \equalref^{\eqref{ssp3_1/3var}} 4\texttt{Var}[x_k],
\end{split}
\end{equation}

\begin{equation}
\label{ssp3_(1)cov}
\begin{split}
    \texttt{Cov}[x_k,x^{(1)}]
    & \equalref^{\eqref{cov_property}} \texttt{Cov}[x_k,x_{k+\frac{1}{3}}] + \texttt{Cov}[x_k,f(x_{x+\frac{1}{3}})] \\
    & \equalref^{(\ref{assumption2},\ref{ssp3_1/3cov})} \texttt{Var}[x_k].
\end{split}
\end{equation}

By using $x^{(1)} = x_{k+\frac{1}{3}}+f(x_{k+\frac{1}{3}})$,
\begin{equation}
\label{ssp3_2/3var}
\begin{split}
    \texttt{Var}[x_{k+\frac{2}{3}}]
    & \equalref^{\eqref{eq:sspblock3}} \texttt{Var}\left[\frac{3}{4}x_k + \frac{1}{4}x_{k+\frac{1}{3}}+\frac{1}{4}f(x_{k+\frac{1}{3}})\right] \\
    & = \texttt{Var}\left[\frac{3}{4}x_k+\frac{1}{4}x^{(1)}\right] \\
    & \equalref^{\eqref{var_property}} \frac{9}{16}\texttt{Var}[x_k] + \frac{1}{16}\texttt{Var}[x^{(1)}] + \frac{3}{8}\texttt{Cov}[x_k,x^{(1)}] \\
    & \equalref^{(\ref{ssp3_(1)var},\ref{ssp3_(1)cov})} \frac{19}{16}\texttt{Var}[x_k],
\end{split}
\end{equation}
and
\begin{equation}
\label{ssp3_2/3cov}
\begin{split}
    \texttt{Cov}[x_k,x_{k+\frac{2}{3}}]
    & = \texttt{Cov}\left[x_k,\frac{3}{4}x_k+\frac{1}{4}x^{(1)}\right] \\
    & \equalref^{\eqref{cov_property}} \frac{3}{4}\texttt{Cov}[x_k,x_k] + \frac{1}{4}\texttt{Cov}[x_k,x^{(1)}] \\
    & \equalref^{\eqref{ssp3_(1)cov}} \texttt{Var}[x_k].
\end{split}
\end{equation}

Similar to previous steps, let $x^{(2)} = x_{k+\frac{2}{3}}+f(x_{k+\frac{2}{3}})$.
Once again, by applying same procedure,
\begin{equation}
\label{ssp3_(2)var}
\begin{split}
    & \texttt{Var}[x^{(2)}] \\
    & \equalref^{\eqref{var_property}} \texttt{Var}[x_{k+\frac{2}{3}}] + \texttt{Var}[f(x_{k+\frac{2}{3}})] + 2\texttt{Cov}[x_{k+\frac{2}{3}},f(x_{k+\frac{2}{3}})] \\
    & \equalref^{\eqref{assumption2}} \texttt{Var}[x_{k+\frac{2}{3}}] + \texttt{Var}[f(x_{k+\frac{2}{3}})] \\
    & \equalref^{\eqref{assumption1}} 2\texttt{Var}[x_{k+\frac{2}{3}}] \\
    & \equalref^{\eqref{ssp3_2/3var}} \frac{19}{8}\texttt{Var}[x_k],
\end{split}
\end{equation}
and
\begin{equation}
\label{ssp3_(2)cov}
\begin{split}
    \texttt{Cov}[x_k,x^{(2)}]
    & \equalref^{\eqref{cov_property}} \texttt{Cov}[x_k,x_{k+\frac{2}{3}}] + \texttt{Cov}[x_k,f(x_{k+\frac{2}{3}})] \\
    & \equalref^{\eqref{assumption2}} \texttt{Cov}[x_k,x_{k+\frac{2}{3}}] \\
    & \equalref^{\eqref{ssp3_2/3cov}} \texttt{Var}[x_k].
\end{split}
\end{equation}

Finally, since $x^{(2)} = x_{k+\frac{2}{3}}+f(x_{k+\frac{2}{3}})$,
\begin{equation*}
\begin{split}
    \texttt{Var}[x_{k+1}]
    & \equalref^{\eqref{eq:sspblock3}} \texttt{Var}\left[\frac{1}{3}x_k + \frac{2}{3}x_{k+\frac{2}{3}} + \frac{2}{3} f(x_{k+\frac{2}{3}})\right] \\
    & = \texttt{Var}\left[\frac{1}{3}x_k + \frac{2}{3}x^{(2)}\right] \\
    & \equalref^{\eqref{var_property}} \frac{1}{9}\texttt{Var}[x_k] + \frac{4}{9}\texttt{Var}[x^{(2)}] + \frac{4}{9}\texttt{Cov}[x_k,x^{(2)}] \\
    & \equalref^{(\ref{ssp3_(2)var},\ref{ssp3_(2)cov})} \frac{1}{9}\texttt{Var}[x_k] + \frac{19}{18}\texttt{Var}[x_k] + \frac{4}{9}\texttt{Var}[x_k] \\
    & = \frac{29}{18}\texttt{Var}[x_k].
\end{split}
\end{equation*}
\section{Comparison non-TVD scheme and TVD scheme}
\label{sec:tvdcompare}
\begin{figure}[t]
    \centering
    \begin{subfigure}[b]{0.5\columnwidth}
        \centering
        \includegraphics[width=\textwidth]{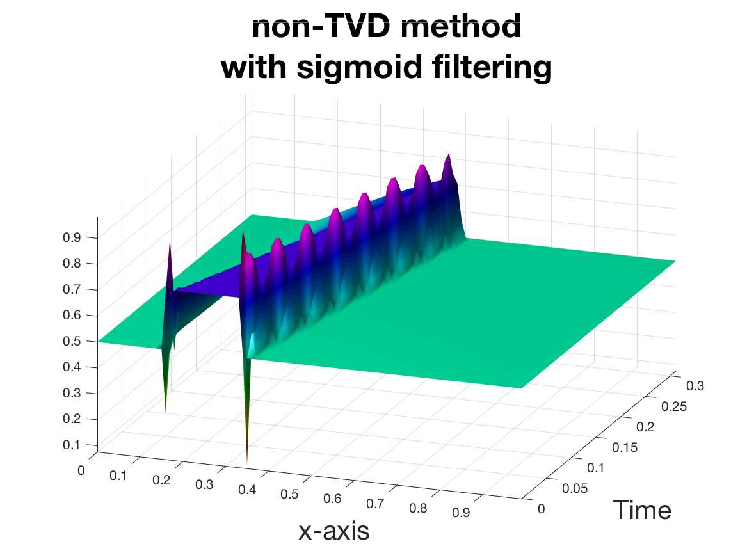}
        \caption{non-TVD scheme}
        \label{subfig:euler}
    \end{subfigure}%
    \begin{subfigure}[b]{0.5\columnwidth}
        \centering
        \includegraphics[width=\textwidth]{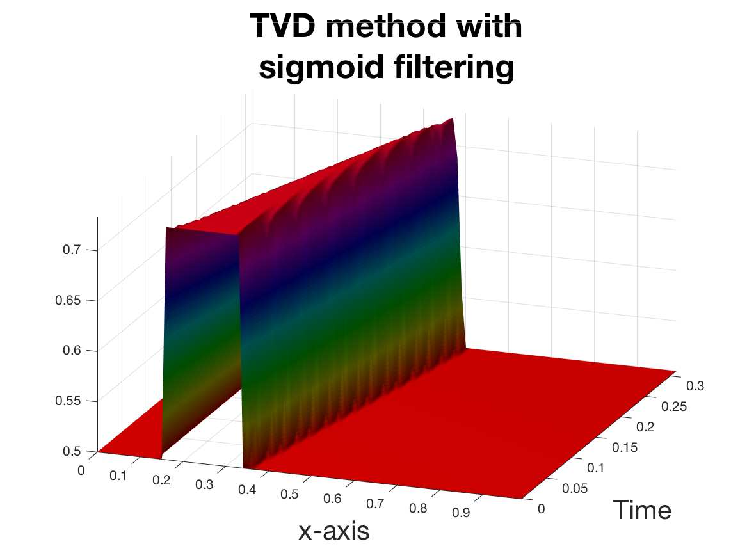}
         \caption{TVD scheme}
        \label{subfig:ssp3}
    \end{subfigure}
    \caption{Two numerical solutions of the inviscid Burgers' equations
    using two different time discretizations are presented above.
    (a) shows the numerical solution with the non-TVD scheme in \eqref{e:nontvd} while (b) is the numerical solution with the SSP3 discretization as in \eqref{eq:ssp3}.
    After computing numerical solutions of the Burgers' equations, the solutions are filtered through the sigmoid function as an activation function.
    Evidently, the left panel (a) displays wild oscillations while the right panel (b) displays accurate numerical solutions.
    }
    \label{fig:title_figure2}
\end{figure}
In Figure \ref{fig:title_figure2}, we implemented numerical solutions of the inviscid Burgers' equations
    \begin{equation}
    \begin{split}
        & u_t + u u_x = 0, \quad x \in (0,1),\\
        & u(0,t) = u(1,t),\\
        & u(x,0) = u_0,
    \end{split}
    \end{equation}
    using two different time discretizations; non-TVD scheme in (a) and TVD scheme in (b).
    The initial condition $u_0(x)$ is 
    \begin{equation*}
        u_0(x) := 
        \begin{cases}
            0,& 0 < x \leq 1/6,\\
            1,& 1/6 < x \leq 2/6,\\
            0,& 2/6 < x < 1.
        \end{cases}
    \end{equation*}
The same equations and initial conditions were also used in Figure 1 of the main manuscript.    
For the spatial discretization, we adopt the third-order weighted essentially non-oscillatory (WENO) schemes.
For numerical computations, the following configurations are used:
\begin{equation*}
\begin{split}
    & N = \text{number of grid points in $x$} = 100,\\
    & h = \text{Time step size} = 0.8/N,\\
    & T = \text{final time} = 0.3.
\end{split}    
\end{equation*}
The left panel (a) shows the numerical solution with non-TVD time discretization of the second order while the right panel (b) presents the numerical solution with the SSP-2 discretization stated in \eqref{eq:ssp2}.
More precisely, in the left panel, we used the second order non-TVD scheme 
    \begin{equation} \label{e:nontvd}
    \begin{split}
        & u^{(1)} = u^n - 20 \Delta t L(u^n),\\
        & u^{n+1} = u^n + \frac{41}{40} \Delta t L(u^n) - \frac{1}{40} \Delta t L(u^{(1)}).
    \end{split}
    \end{equation}
After computing numerical solutions of the Burgers' equations, the solutions are filtered through the sigmoid function as activation function.

\section{Adversarial Training Details}
\label{sec:adversarial_training}
\begin{algorithm}[t]
\caption{Projected Gradient Descent}
\label{alg:pgd}
\textbf{Input: }{Clean image $x_{\text{nat}}\in[0,1]^m$ and corresponding label $y$, model $h_{\theta}$, loss function $\ell$, step $\alpha$, bound $\epsilon$, \# of iteration $K$}, metric $p$. \\
\textbf{Output: }{Candidate adversarial example $x_{\text{adv}}$.}
\begin{algorithmic}
\State $x_{\text{adv}}:=x_{\text{nat}}$
\State $\texttt{feasible-set} = \{x'|\|x'-x_{\text{nat}}\|_p \le \epsilon \}\cap[0,1]^m$
\For{i \textbf{in} range($K$)}
\State $x_{\text{adv}} = x_{\text{adv}} + \alpha\cdot\texttt{sign}(\nabla_{x_{\text{adv}}} \ell (h_{\theta}(x_{\text{adv}}),y))$
\State $x_{\text{adv}}=\Pi(x_{\text{adv}}, \texttt{feasible-set})$
\EndFor

\State \textbf{return} $x_{\text{adv}}$
\end{algorithmic}
\end{algorithm}
\begin{algorithm}[t]
\caption{PGD Adversarial Training}
\label{alg:advtrain}
\textbf{Input: }{Training data minibatch $(x_i,y_i)$ with $i\in\{1,\cdots,N\}$, initialized model $h_{\theta}$, training epochs $K$, PGD attack algorithm \texttt{PGD}, Optimization method \texttt{optim}}. \\
\textbf{Output: }{Trained model $h_{\theta}$}
\begin{algorithmic}
\For{i \textbf{in} range($K$)}
\For{j \textbf{in} range($N$)}
\State Sample random $\delta \in Uniform(-\epsilon,\epsilon)$
\State $x_{j,\text{adv}} = x_{j,\text{nat}} + \delta$
\State $x_{j,\text{adv}} = \texttt{PGD}(x_{j,\text{adv}},x_{j,\text{nat}},y_j)$
\State $\theta := \texttt{optim}( h_{\theta}(x_{j,\text{adv}}), y_j)$
\EndFor
\EndFor

\State \textbf{return} $h_{\theta}$
\end{algorithmic}
\end{algorithm}

In this section, we briefly introduce the adversarial training which we used in our experiments.

Adversarial training is state-of-the-art methodology for defending adversarial attacks~\cite{madry2017towards,wang2019on,xie2019feature}.
As we mentioned in our main paper, the objective of adversarial training is to minimize the adversarial risk given as
\begin{equation}
\label{eq:adv_risk}
    R_{adv}(h_\theta) = \mathbb{E}_{(x,y)\sim D}\big[\max\limits_{\delta \in \Delta} \mathcal{L}(h_\theta (x+\delta),y)\big],
\end{equation}
where the all notations are the same as our main paper.
Strictly speaking, the set $\Delta$ has finite number of elements, so the exact solutions exist which maximize the loss $\mathcal{L}(h_{\theta}(x),y)$.
However, as we mentioned in our main paper, numerically solving this problem is intractable.
Therefore, most of works estimate the $\max\limits_{\delta \in \Delta} \mathcal{L}(h_{\theta}(x),y)$ by using PGD; see Algorithm \ref{alg:pgd}.
Further, the randomness is injected during adversarial training, and this may help the robustness~\cite{madry2017towards,xie2019feature}.
The description of adversarial training is shown in Algorithm \ref{alg:advtrain}.
\section{Exploratory Network Analysis}
\label{sec:netanalysis}
\begin{table}[t]
\centering
\resizebox{\columnwidth}{!}{
\begin{tabular}{cccccccccc}
\toprule
$\epsilon$ &      Natural&      1 &      2 &      3 &      4 &      5 &      6 &      7 &      8 \\
\midrule
ResNet  &  \textbf{88.14} &  83.00 &  77.22 &  70.49 &  64.46 &  58.45 &  52.50 &  47.21 &  42.29 \\
SSP-2 &  87.59 &  83.04 &  77.61 &  71.77 &  66.09 &  60.27 &  54.68 &  49.08 &  44.73 \\
SSP-3 &  87.51 &  \textbf{83.31} &  \textbf{78.49} &  \textbf{72.70} &  \textbf{66.61} &  \textbf{60.90} &  \textbf{55.28} &  \textbf{50.20} &  \textbf{45.40} \\
\bottomrule
\end{tabular}
}
\caption{Network performance against FGSM adversarial attacks when trained with PGD training ($\alpha=1/255$). SSP-3 is more robust than ResNet and SSP-2; approximately 3\%.}
\label{tab:result_fgsm}
\end{table}

\begin{table}[t]
\centering
\resizebox{\columnwidth}{!}{
\begin{tabular}{cccccccccc}
\toprule
{$\epsilon$} &      Natural &      1 &      2 &      3 &      4 &      5 &      6 &      7 &      8 \\
\midrule
ResNet  &  \textbf{88.14} &  82.74 &  76.00 &  67.69 &  59.33 &  51.01 &  43.25 &  36.96 &  31.31 \\
SSP-2 &  87.59 &  82.90 &  76.87 &  70.06 &  62.59 &  54.83 &  47.35 &  41.00 &  35.02 \\
SSP-3 &  87.51 &  \textbf{83.16} &  \textbf{77.78} &  \textbf{70.75} &  \textbf{63.19} &  \textbf{55.53} &  \textbf{48.51} &  \textbf{42.08} &  \textbf{36.13}\\
\bottomrule
\end{tabular}
}
\caption{Network performance against PGD adversarial attacks when trained with PGD training ($\alpha=1/255$). SSP-3 is more robust than ResNet and SSP-2; approximately 5\%.}
\label{tab:result_pgd}
\end{table}
\begin{table}[t]
    \centering
    \begin{tabular}{c|c|c|c}
        \toprule
        Model & Clean & FGSM & PGD$_{20}$ \\
        \midrule\midrule
        ResNet & 0.9090 & 0.8562    & 0.8179 \\
        SSP-2 & 0.9132  & 0.8591    & 0.8252 \\
        SSP-3 & 0.9110  & \textbf{0.8639}    & \textbf{0.8295} \\
        SSP-adap & 0.9098    & 0.8621    & 0.8264 \\
        \bottomrule
    \end{tabular}
    \caption{
    Result on Fashion-MNIST. All the models follow the same architecture which used in MNIST experiment in main paper. The number of blocks is 20, with using Group Normalization.
    We perform adversarial training with $\epsilon=0.1$, $\alpha=0.02$ with 10 iterations.
    For evaluating robustness, $\alpha=0.01$ with 20 iterations are used in PGD attack (PGD$_{20}$).
    }
    \label{tab:fmnist}
\end{table}
\begin{table}[t]
    \centering
    \begin{tabular}{c|c|c}
        \toprule
        Model & Clean & PGD$_{20}$ \\
        \midrule\midrule
        ResNet & 0.4648 & 0.1738 \\
        SSP-2 & 0.4386  & 0.1761 \\
        SSP-3 & 0.4529  & \textbf{0.1955} \\
        \bottomrule
    \end{tabular}
    \caption{
    Result on Tiny-Imagenet. We perform adversarial training with $\epsilon=8/255$, $\alpha=2/255$ with 5 iterations. For evaluating robustness, $\alpha=2/255$ with 20 iterations are used in PGD attack (PGD$_{20}$).
    }
    \label{tab:tiny}
\end{table}

\textcolor{black}{In this section, we provide more experimental data on the CIFAR-10 dataset; as well as other well known datasets: Fashion-MNIST and Tiny-Imagenet.}

We have trained ResNet, SSP-2 and SSP-3 on the CIFAR10 dataset, with $N=5,~K=7$, and PGD adversarial training with $\alpha=1/255$; in order to gauge performance and robustness of our architecture. 
We were able to perform various attacks on the network using the PGD and FGSM methods. 
We believe that this can be optimized with higher order methods, specifically SSP.
Intuitively speaking, we are expecting performance to increase as a result of using a more accurate approximation.
Also, accuracy should increase as we increase the order of the numerical approximation.

We noticed that when all three models were attacked via FGSM, or PGD, with $\alpha=1/255$, and various $\epsilon$, that the higher order methods outperformed ResNet from roughly $3\sim5\%$. 
We observed that SSP-3 outperforms SSP-2, which outperforms ResNet; consistently. 
Furthermore, as the strength of the perturbation was increased, SSP networks became more resilient to adversarial attacks than ResNet.
This results in behavior that is truly characteristic of numerical methods, in that accuracy is increased as higher order methods are implemented, and stability is preserved.

What is perhaps the most notable is that SSP-3 is extremely more robust than ResNet whether it is attacked via FGSM or PGD.
Robustness is achieved without introducing more parameters, or a dramatic increase to computational power. 
Our architecture achieves comparable performance on unperturbed images and superior performance with respect to adversarial attacks. 


\textcolor{black}{
Next, we evaluate the robustness on Fashion-MNIST~\cite{xiao2017fashion} dataset.
The architecture of model is same as the model used in MNIST, but the only difference is the number of blocks and maximum perturbation range ($\epsilon$).
The results are shown in Table \ref{tab:fmnist} and are consistent with our previous assumptions and results.
}

\textcolor{black}{
Last but not least, we conduct an experiment on the more challenging dataset, Tiny-Imagenet \cite{le2015tiny}.
The models used in Tiny-Imagenet experiment are composed of 4 groups of blocks and each group has 10 blocks.
Table \ref{tab:tiny} shows the top-1 accuracy of natural samples and adversarial examples generated by PGD attack.
As the result shows, all the SSP networks show better robustness than ResNet.
}
\section{Suppression on Perturbation Growth}
\label{sec:suppression}
\captionsetup[sub]{font=small,labelfont={bf,sf}}
\begin{figure*}[h]
    \centering
    \begin{subfigure}{0.245\textwidth}
        \centering
        \includegraphics[width=\textwidth]{figures/pgr_6_7_norm1.PNG}
        \caption{N$=$6, K$=$7, p$=$1}
    \end{subfigure}%
    \begin{subfigure}{0.245\textwidth}
        \centering
        \includegraphics[width=\textwidth]{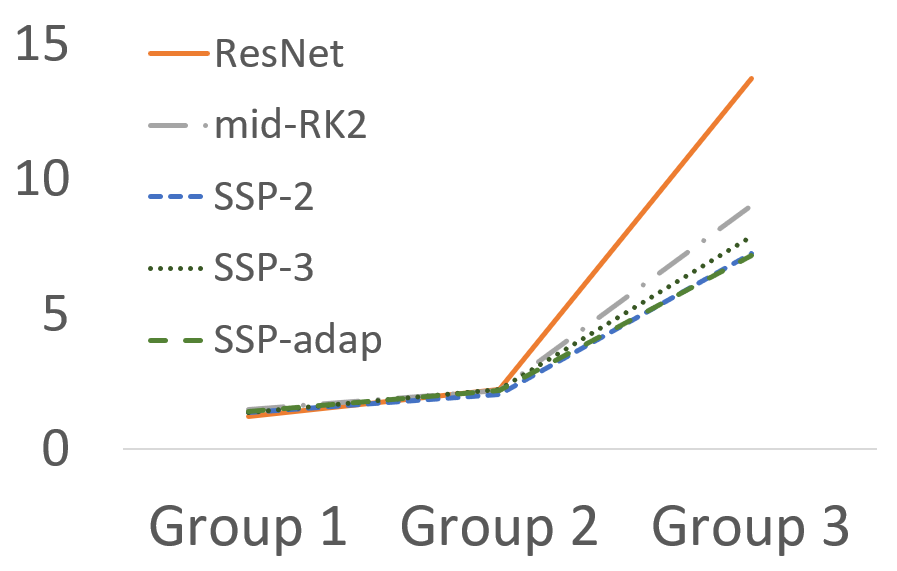}
        \caption{N$=$6, K$=$12, p$=$1}
    \end{subfigure}%
    \begin{subfigure}{0.245\textwidth}
        \centering
        \includegraphics[width=\textwidth]{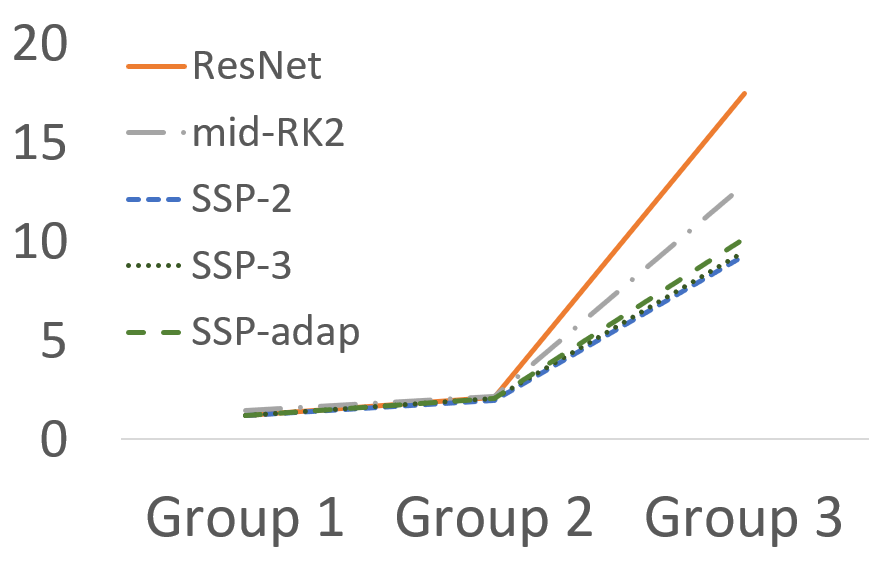}
        \caption{N$=$10, K$=$7, p$=$1}
    \end{subfigure}%
    \begin{subfigure}{0.245\textwidth}
        \centering
        \includegraphics[width=\textwidth]{figures/pgr_10_12_norm1.PNG}
        \caption{N$=$10, K$=$12, p$=$1}
    \end{subfigure}%
    \qquad
     \begin{subfigure}{0.245\textwidth}
        \centering
        \includegraphics[width=\textwidth]{figures/pgr_6_7_norm2.PNG}
        \caption{N$=$6, K$=$7, p$=$2}
    \end{subfigure}%
    \begin{subfigure}{0.245\textwidth}
        \centering
        \includegraphics[width=\textwidth]{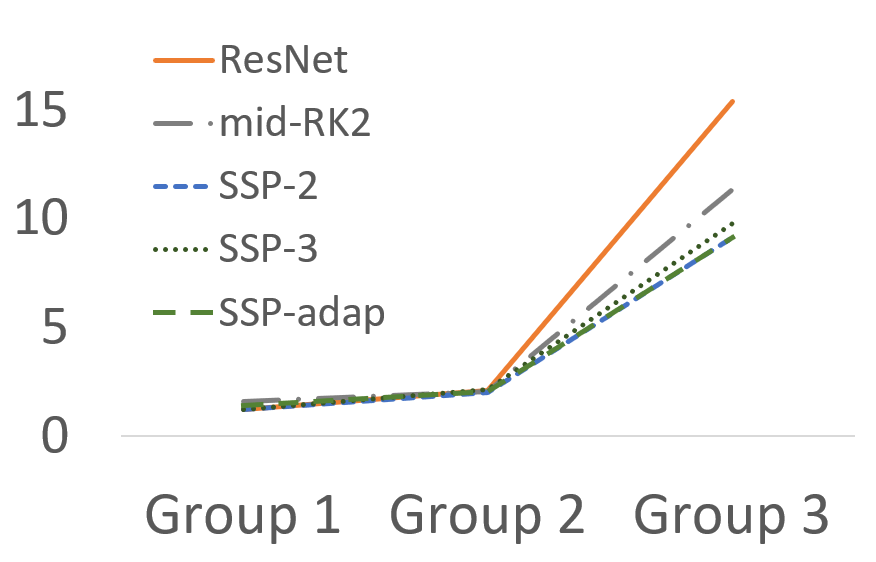}
        \caption{N$=$6, K$=$12, p$=$2}
    \end{subfigure}%
    \begin{subfigure}{0.245\textwidth}
        \centering
        \includegraphics[width=\textwidth]{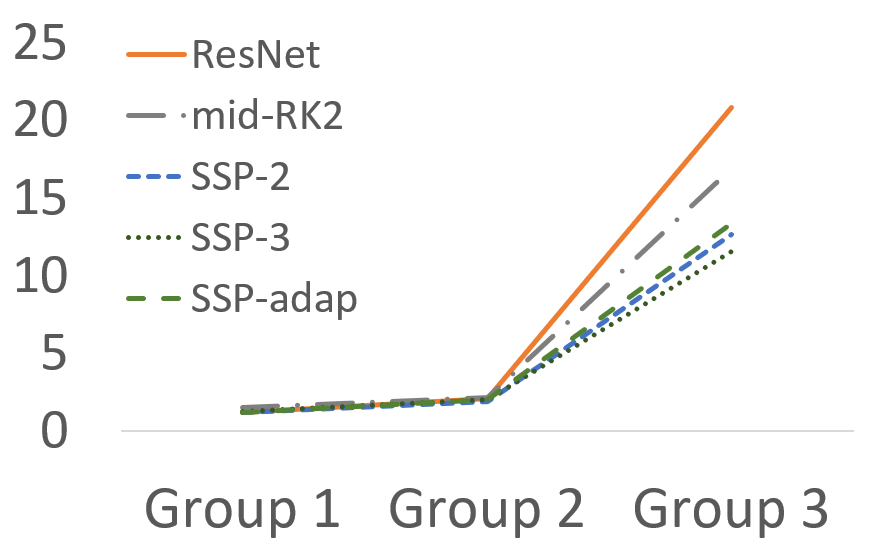}
        \caption{N$=$10, K$=$7, p$=$2}
    \end{subfigure}%
    \begin{subfigure}{0.245\textwidth}
        \centering
        \includegraphics[width=\textwidth]{figures/pgr_10_12_norm2.PNG}
        \caption{N$=$10, K$=$12, p$=$2}
    \end{subfigure}%
    \caption{
    Perturbation growth ratio of clean samples and its adversarial counterparts. As the perturbation evolves through networks, SSPNets have lower ratio than ResNet.
    }
    \label{fig:normratio_random}
\end{figure*}


In this section, we present the perturbation growth ratio of all the networks used in the CIFAR-10 experiments.
Recall that the perturbation growth ratio is given by
\begin{equation*}
    \texttt{PGR}(f)=\mathbb{E}_{x\sim\mathcal{D}}\left[\mathbb{E}_{x'\sim\mathcal{X'}}\left[\frac{\|f(x) - f(x')\|_p}{\|x - x'\|_p}\right]\right],\quad p \in \{1,2\},
\end{equation*}
where each corrupted sample $x'$ is sampled from a small neighborhood of $x$, i.e., $\mathcal{X'}$, and $p$ defines a type of norm either $\ell_1$ or $\ell_2$.

In Figure \ref{fig:normratio_random}, all the corrupted sample $x'$ is the adversarial example generated by PGD attack with 20 iterations for each model.
Despite there is no other regularization using Lipschitzness or Jacobian, all the SSPNets have lower perturbation growth ratio than ResNet.

\end{document}